\documentclass{article}

\usepackage{microtype}
\usepackage{graphicx}
\usepackage{subfigure}
\usepackage{booktabs} 

\usepackage{hyperref}

\usepackage[accepted]{icml2020-arxiv}

\usepackage{bm}
\usepackage{amsthm}
\usepackage{amsmath}
\usepackage{amssymb}
\usepackage{amsfonts}
\usepackage{bbm}
\usepackage{dsfont}

\DeclareMathOperator*{\argmax}{arg\,max}

\usepackage{xcolor}

\newcommand{\Jac}{\bm{J_y}}
\newcommand{\thetab}{\bm{\theta}}
\newcommand{\dbf}{\mathbf{d}}
\newcommand{\ybf}{\mathbf{y}}

\newcommand{\bound}{\widehat{I}(\dbf, {\bm{\psi}})}
\newcommand{\NN}{T_{\bm{\psi}}(\thetab, \ybf)}
\newcommand{\Ej}{\mathbb{E}_{p(\thetab, \ybf \mid \dbf)}}
\newcommand{\Em}{\mathbb{E}_{p(\thetab) p(\ybf \mid \dbf)}}
\newcommand{\psib}{\bm{\psi}}

\icmltitlerunning{BED for Implicit Models using MINE}

\begin{document}

\twocolumn[
\icmltitle{Bayesian Experimental Design for Implicit Models \\ by Mutual Information Neural Estimation}



\icmlsetsymbol{equal}{*}

\begin{icmlauthorlist}
\icmlauthor{Steven Kleinegesse}{ed}
\icmlauthor{Michael U. Gutmann}{ed}
\end{icmlauthorlist}

\icmlaffiliation{ed}{School of Informatics, University of Edinburgh, Edinburgh, UK}

\icmlcorrespondingauthor{Steven Kleinegesse}{steven.kleinegesse@ed.ac.uk}

\icmlkeywords{Machine Learning, ICML}

\vskip 0.3in
]



\printAffiliationsAndNotice{}  

\begin{abstract}
Implicit stochastic models, where the data-generation distribution is intractable but sampling is possible, are ubiquitous in the natural sciences. The models typically have free parameters that need to be inferred from data collected in scientific experiments. A fundamental question is how to design the experiments so that the collected data are most useful. The field of Bayesian experimental design advocates that, ideally, we should choose designs that maximise the mutual information (MI) between the data and the parameters. For implicit models, however, this approach is severely hampered by the high computational cost of computing posteriors and maximising MI, in particular when we have more than a handful of design variables to optimise. In this paper, we propose a new approach to Bayesian experimental design for implicit models that leverages recent advances in neural MI estimation to deal with these issues. We show that training a neural network to maximise a lower bound on MI allows us to jointly determine the optimal design and the posterior. Simulation studies illustrate that this gracefully extends Bayesian experimental design for implicit models to higher design dimensions.
\end{abstract}

\section{Introduction} \label{sec:introduction}  

Many processes in nature can be described by a parametric statistical model from which we can simulate data. When the corresponding data-generating distribution is intractable we refer to it as an implicit model. These models are abundant in science and engineering, having been used, for instance, in high-energy physics~\citep{Agostinelli2003}, cosmology~\citep{Schafer2012}, epidemiology~\citep{Corander2017}, cell biology~\citep{Vo2015} and pharmacokinetics~\citep{Donnet2013}. Usually, we wish to infer the free parameters $\thetab$ of the implicit model in order to understand the underlying natural process better or to predict some future events. Since the likelihood function is intractable for implicit models, we have to revert to likelihood-free inference methods such as approximate Bayesian computation \citep[for recent reviews see e.g.\ ][]{Lintusaari2017, Sisson2018}.

While considerable research effort has focused on developing efficient likelihood-free inference methods \citep[e.g.\ ][]{Papamakarios2019, Chen2019, Gutmann2016, Papamakarios2016, Ong2018}, the quality of the estimated parameters $\thetab$ ultimately depends on the quality of the data $\ybf$ that are available for inference in the first place. We here consider the scenario where we have control over experimental designs $\dbf$ that affect the data collection process. For example, these might be the spatial location or time at which we take measurements, or they might be the stimulus that is used to perturb the natural system. Our overall goal is then to find experimental designs $\dbf$ that yield the most information about the model parameters.

In Bayesian experimental design (BED), we construct and optimise a utility function $U(\dbf)$ that indicates the value of a design $\dbf$. A choice for the utility function that is firmly rooted in information theory is the mutual information (MI) $I(\thetab, \ybf ; \dbf)$ between model parameters and data, 
\begin{equation} \label{eq:mi}
  I(\thetab, \ybf ; \dbf) = \mathbb{E}_{p(\thetab, \ybf \mid \dbf)} \left[ \log{\frac{p(\thetab \mid \ybf, \dbf)}{p(\thetab)}} \right],
\end{equation}
where $p(\thetab \mid \ybf, \dbf)$ is the posterior distribution of $\thetab$ given data $\ybf$ that were obtained with design $\dbf$, and $p(\thetab)$ is the prior belief of $\thetab$.\footnote{It is typically assumed that $p(\thetab)$ is not affected by $\dbf$.} MI describes the expected uncertainty reduction in the model parameters $\thetab$ when collecting data with experimental design $\dbf$, which makes it an effective utility function for BED. Since MI depends on the full posterior $p(\thetab \mid \ybf, \dbf)$, it is sensitive to deviations from the prior, e.g.\ nonlinear correlations or multi-modality, that other utility functions do not capture~\citep{Ryan2016}.

Although a quantity with highly desirable properties, MI is notoriously difficult to compute and maximise. The key problems in the context of implicit models are 1) that the posteriors are hard to estimate and that obtaining samples from them is expensive as well; 2) that the functional relationship between the designs and the MI is unknown and that (approximate) gradients are generally not available either.

The first problem made the community turn to approximate Bayesian computation and less powerful utility functions, e.g.\ based on the posterior variance \citep{Drovandi2013}, or approximations with kernel density estimation \citep{Price2016}. More recently, \citet{Kleinegesse2019} have shown that likelihood-free inference by density ratio estimation \citep{Thomas2016} can be used to estimate the posterior and the mutual information at the same time, which alleviates the first problem to some extent. The second problem has been addressed by using gradient-free optimisation techniques such as grid-search, sampling \citep{Muller1999}, evolutionary algorithms \citep{Price2018}, Gaussian-process surrogate modelling \citep{Overstall2018} and Bayesian optimisation \citep{Kleinegesse2019}. However, these approaches generally do not scale well with the dimensionality of the design variables $\dbf$~\citep[e.g.][]{Spall2005}.

In BED, our primary goal is to find the designs $\dbf$ that maximise the MI and not estimating the MI to a high accuracy. Rather than spending resources on estimating MI accurately for non-optimal designs, a potentially more cost-effective approach is thus to relax the problem and to determine the designs that maximise a lower bound of the MI instead, while tightening the bound at the same time. In our paper, we take this approach and show that for a large class of implicit models, the lower bound can be tightened and maximised by gradient ascent, which addresses the aforementioned scalability issues. Our approach leverages the Mutual Information Neural Estimation (MINE) method of~\citet{Belghazi2018} to perform BED --- we thus call it MINEBED.\footnote{The research code for this paper is available at: \url{https://github.com/stevenkleinegesse/minebed}} We show later on that in addition to gradient-based experimental design, MINEBED also provides us with an approximation of the posterior $p(\thetab \mid \ybf, \dbf)$ so that no separate and oftentimes expensive likelihood-free inference step is needed.

\paragraph{Related Work}\citet{Foster2019} have recently considered the use of MI lower bounds for experimental design. Their approach is based on variational approximations to the posterior and likelihood. While the authors note that this introduces a bias, in practice, this may actually be acceptable. However, the optimal designs were determined by Bayesian optimisation and, like in the aforementioned previous work, this approach may thus also suffer from scalability problems. The authors rectify this in a follow-up paper~\citep{Foster2019sgd} but their experiments focus on the explicit setting where likelihood functions are tractable or approximations are available, unlike our paper. In theory, their method is also applicable in the likelihood-free setting.


\section{The MINEBED Method} \label{sec:minebed}

We here show how to perform experimental design for implicit models by maximising a lower bound of the mutual information (MI) between model parameters $\thetab$ and data $\ybf$. The main properties of our approach are that for a large and interesting class of implicit models, the designs can be found by gradient ascent and that the method simultaneously yields an estimate of the posterior.

We start with the $\text{MINE-}f$ MI lower bound of~\citet{Belghazi2018}, which we shall call $\bound$, 
\begin{equation} \label{eq:minebed}
\begin{split}
\bound = \, &\Ej \left[\NN \right] \\
		&- e^{-1}\Em\left[e^{\NN}\right],
\end{split}
\end{equation}
where $\NN$ is a neural network parametrised by ${{\psib}}$, with $\thetab$ and $\ybf$ as input. This is also the lower bound of~\citet{Nguyen2010} and the $f$-GAN KL of~\citet{Nowozin2016}. We note that while we focus on the above lower bound, our approach could also be applied to other MI lower bounds of e.g.~\citet{Poole2019}, \citet{Foster2019} and \citet{Foster2019sgd}. Moreover, comparing the performance of different MI lower bounds is not the aim of this paper and we do not claim that the particular bound in~\eqref{eq:minebed} is superior to others.

Importantly,~\citet{Belghazi2018} showed that the bound in \eqref{eq:minebed} can be tightened by maximising it with respect to the neural network parameters ${{\psib}}$ by gradient ascent and that for flexible enough neural networks we obtain $I(\thetab, \ybf ; \dbf) = \max_{{{\psib}}} \bound $. Our experimental design problem can thus be formulated as 
\begin{equation} \label{eq:opt}
\dbf^\ast = \argmax_{\dbf} \max_{{{\psib}}} \left\{\bound \right\}.
\end{equation}
The main difficulty is that for implicit models, the expectations in the definition of $\bound$ generally depend in some complex manner on the design variables $\dbf$, complicating gradient-based optimisation.

\subsection{Experimental Design with Gradients}

We are interested in computing the derivative $\nabla_{\dbf} \bound$, 
\begin{align}
\nabla_{\dbf} \bound &= \nabla_{\dbf} \Ej \left[\NN \right] \nonumber\\
		&- \nabla_{\dbf} e^{-1}\Em \left[e^{\NN}\right]. 
\label{eq:grad_issue}
\end{align}
For implicit models we do not have access to the gradients of the joint distribution $p(\thetab, \ybf | \dbf)$ and marginal distribution $p(\ybf | \dbf)$. Thus, we cannot simply pull $\nabla_{\dbf}$ into the integrals defining the expectations and compute the derivatives. For the same reason we cannot use score-function estimators~\citep{Mohamed2019}, as they rely on the analytic derivative of the log densities.

The alternative pathwise gradient estimators~\citep[see][for a review]{Mohamed2019}, however, are well suited for implicit models. They require that we can sample from the data-generating distribution $p(\ybf | \thetab, \dbf)$ by sampling from a base distribution $p(\bm{\epsilon})$ and then transforming the samples through a (nonlinear) deterministic function $\mathbf{h}(\bm{\epsilon}; \thetab, \dbf)$ called the sampling path ~\citep{Mohamed2019}, i.e.
\begin{equation} \label{eq:path}
\ybf \sim p(\ybf | \thetab, \dbf) \iff \ybf = \mathbf{h}(\bm{\epsilon}; \thetab, \dbf), \quad \bm{\epsilon} \sim p(\bm{\epsilon}).
\end{equation}
For implicit models, this requirement is automatically satisfied because the models are specified by such transformations in the first place.

We first factorise the joint $p(\thetab, \ybf | \dbf) = p(\ybf | \thetab, \dbf) p(\thetab)$, where we again assume that $p(\thetab)$ is not affected by $\dbf$. Specifying $p(\ybf | \thetab, \dbf)$ implicitly in terms of the sampling path $\mathbf{h}(\bm{\epsilon}; \thetab, \dbf)$ then allows us to invoke the law of the unconscious statistician~\citep[e.g.][]{Grimmett2001} and, for instance, rewrite the first expectation in~\eqref{eq:minebed} as
\begin{equation}
\Ej \left[ \NN \right] = \mathbb{E}_{p(\thetab)p(\bm{\epsilon})} \left[ T_{{\psib}}(\thetab, \mathbf{h}(\bm{\epsilon}; \thetab, \dbf)) \right],
\end{equation}
%
Since neither $p(\bm{\epsilon})$ nor $p(\thetab)$ directly depend on $\dbf$, taking the derivative of the above expectation, which proved problematic before, yields 
\begin{equation}
\nabla_{\dbf} \Ej \left[ \NN \right] = \mathbb{E}_{p(\thetab)p(\bm{\epsilon})} \left[ \nabla_{\dbf} T_{{\psib}}(\thetab, \mathbf{h}(\bm{\epsilon}; \thetab, \dbf)) \right].
\end{equation}
We can compute the derivative of the neural network output with respect to designs via the chain rule, i.e.\
\begin{equation} \label{eq:nn_gradd}
\nabla_{\dbf} T_{{\psib}}(\thetab, \mathbf{h}(\bm{\epsilon}; \thetab, \dbf)) = \Jac^\top \nabla_{\ybf} T_{{\psib}}(\thetab, \ybf) \rvert_{\ybf = \mathbf{h}(\bm{\epsilon}; \thetab, \dbf)},
\end{equation}
where $\Jac$ is the Jacobian matrix containing the derivatives of the
vector-valued $\ybf=\mathbf{h}(\bm{\epsilon}; \thetab, \dbf)$ with
respect to $\dbf$\footnote{We use the definition that ${\Jac}_{ij} = \partial y_i / \partial d_j$.} and $\nabla_{\ybf} T_{{\psib}}(\thetab, \ybf)$ is the
gradient of the neural network $\NN$ with respect to its inputs
$\ybf$, which is provided by most machine-learning libraries via auto-differentiation. The
computation of the Jacobian $\Jac$, or the Jacobian-gradient product, is
the main technical obstacle, and below we discuss in which cases we
can compute it and what to do when we cannot.

For the second expectation in~\eqref{eq:grad_issue}, which is defined over the marginal distribution, we have to apply pathwise gradient estimators in a similar manner to above. However, we only know the sampling path of the data-generating distribution, and not the sampling path of the marginal random variable $\ybf \sim p(\ybf \mid \dbf)$. Thus, we need to re-write the marginal distribution as $p(\ybf \mid \dbf) = \mathbb{E}_{p(\thetab^\prime)}\left[ p(\ybf \mid \thetab^\prime, \dbf) \right]$, where $\thetab^\prime$ follows the same distribution as $\thetab$, which then allows us to follow the procedure as described above.

This gives us the following pathwise gradient estimator of the MI lower bound 
\begin{multline} \label{eq:minebed_gradient}
\nabla_{\dbf}\bound = \mathbb{E}_{p(\thetab)p(\bm{\epsilon})}\left[\nabla_{\dbf}\NN \right] \\
- e^{-1}\mathbb{E}_{p(\thetab)p(\bm{\epsilon})p(\thetab^{\prime})}\left[e^{T_{{\psib}}(\thetab, \ybf^\prime)} \nabla_{\dbf}T_{{\psib}}(\thetab^{\prime}, \ybf^\prime)\right],
\end{multline}
where $\ybf = \mathbf{h}(\bm{\epsilon}; \thetab, \dbf)$, $\ybf^\prime
= \mathbf{h}(\bm{\epsilon}; \thetab^\prime, \dbf)$ and the gradients
of $\NN$ with respect to $\dbf$ are given
in~\eqref{eq:nn_gradd}. Throughout this work, we approximate all
expectations, i.e.~in the MI lower bound in~\eqref{eq:minebed} and in
the corresponding gradients in~\eqref{eq:minebed_gradient}, via a
sample-average.

With the gradient in \eqref{eq:minebed_gradient} in hand, we can maximise the MI lower bound in~\eqref{eq:minebed} with respect to both the neural network parameters ${{\psib}}$ and the experimental designs $\dbf$ jointly by gradient ascent. Gradient-based optimisation is key to scaling experimental design to higher dimensions and the joint maximisation tightens the lower bound automatically as the (locally) optimal designs are determined more accurately.

The above development assumes that we can compute the Jacobian $\Jac$
or the Jacobian-gradient product in \eqref{eq:nn_gradd}. While
generally not guaranteed to be computable for implicit models, we do
have access to the exact Jacobian in a number of important cases:
Interestingly, for several implicit models considered in the
literature, the data generating distribution $p(\ybf | \thetab, \dbf)$
might be intractable, but $\mathbf{h}(\bm{\epsilon}; \thetab, \dbf)$
can be expressed analytically in closed form and derivatives with
respect to $\dbf$ are readily available. Examples include the noisy
linear model and the pharmacokinetic model considered in this work.

Another broad class of implicit models where the Jacobian is exactly
computable is the case where the generative process separates into a
black-box latent process that does not depend on $\dbf$ and a
differentiable observation process that is governed or affected by
$\dbf$. In many cases, the latent process is described by stochastic
differential equations and the design variables $\dbf$ are the
measurement times. Whenever we have access to the differential
equations, which we most often have when solving them numerically, the
Jacobian can be computed exactly too. The pharmacokinetic model can be
considered an example of this category as well.

Finally, if the implicit model is written in a differential
programming framework, the required Jacobian-gradient product
in~\eqref{eq:nn_gradd} can be computed via their built-in automatic
differentiation functionality. Given the usefulness of automatic
differentiation packages, we expect their usage only to increase and hence
also the range of models that our method can be applied to.

\subsection{Experimental Design without Gradients}
\label{sec:bo}

We have seen that, for a large class of implicit models, the
optimisation problem in \eqref{eq:opt} can be solved by gradient
ascent on the design variables $\dbf$. However, there might be cases
of implict models and design problems for which the required gradient
can neither be computed exactly nor be reasonably approximated. For
example, when the implicit model is described by a nonlinear
stochastic differential equation and the experimental design is
concerned with interventions and not measurements as assumed above,
the gradient cannot be computed exactly.

However, we can fall back to gradient-free methods to solve the
optimisation problem in \eqref{eq:opt}, at the expense of reduced
scalability. We next describe how Bayesian optimisation
(BO)~\citep[see][for a review]{Shahriari2016} can be used to learn the
optimal design $\dbf^\ast$ when gradients are not available. We
decided to use BO as it smoothes out Monte-Carlo error and has been
shown to perform well in practice.

In brief, BO works by defining a probabilistic surrogate model of the function to be optimised. For the surrogate model we will be using a Gaussian process (GP)~\citep{Rasmussen2005} with a Mat\'ern-5/2 kernel. Based on this surrogate model we have to compute and optimise an acquisition function over the domain, which tells us where to evaluate the function next. We will be using expected improvement (EI), as this has been shown to generally be a cost-effective choice~\citep{Shahriari2016}.

Using an initial design $\dbf_0$, we train the neural network $T_{{{\psib}}}(\thetab, \ybf)$ by gradient ascent on ${{\psib}}$. We then use the obtained parameters ${{\psib}}^\ast$ to compute the lower bound $\widehat{I}(\dbf_0, {{\psib}}^\ast)$, update our probabilistic surrogate model given the computed value, and use the acquisition function to decide for which design $\dbf_1$ to evaluate the lower bound next. For this evaluation, we here opt to re-train the neural network. This is because BO often involves significant exploration and thus may lead to design updates that are vastly different from previous designs. This will change the data distributions significantly enough that the neural network has trouble re-training its parameters, a common issue in transfer learning~\citep{Pan2010}. This procedure is repeated until we have converged to the optimal design $\dbf^\ast$.

We note that BO has been used in BED for implicit models before~\citep[e.g.][]{Foster2019, Kleinegesse2019, Kleinegesse2020}, as it is a popular black-box optimisation tool. Furthermore,~\citet{Foster2019} have used BO to optimise a parametrised MI lower bound, based on the Donsker-Varadhan representation of mutual information~\citep{Donsker1983}. We would like to emphasise that we only present the use of BO for MINEBED as a fall-back method in cases where the required Jacobian cannot be computed.

\subsection{Estimating the Posterior Distribution}

It has been shown that the learned neural network $T_{{{\psib}}^\ast}(\thetab, \ybf)$ is related to the posterior and prior distribution when the bound is tight~\citep{Nguyen2010, Belghazi2018, Poole2019}, 
\begin{align}
p(\thetab \mid \ybf, \dbf) &= e^{T_{{{\psib}}^\ast}(\thetab, \ybf) - 1} p(\thetab). \label{eq:post_density}
\end{align}
This relationship allows us obtain an estimate of the posterior $p(\thetab \mid \ybf, \dbf^*)$ without any additional work after we have determined the optimal design $\dbf^*$ by maximisation of the MI lower bound. Since sampling from the prior $p(\thetab)$ is often easily possible, the above expression further allows us to cheaply obtain posterior samples $\thetab^{(i)} \sim p(\thetab \mid \ybf, \dbf^*)$ in an amortised way for any data $\ybf$ by re-weighing samples from the prior and then using categorical sampling. In practice, any MCMC sampling scheme could be used in place as well, since we are able to quickly evaluate the posterior density via~\eqref{eq:post_density}.

\section{Experiments}

We here showcase our approach to BED for implicit models. First, to
demonstrate scalability, we use gradient ascent to find optimal
designs for a linear model with several noise sources and a
pharmacokinetic model~\citep{Ryan2014}. Then we design experiments to
locate a gas leak in a fluid simulation experiment~\citep{Asenov2019}
where we do not have sampling path gradients. An important property of
mutual information is that it is sensitive to multi-modal posteriors,
unlike previous popular approaches. We verify this in the
supplementary material using a multi-modal, oscillatory toy model.

\subsection{Noisy Linear Model} 

We first consider the classical model where it is assumed that a response variable $y$ has a linear relationship with an experimental design $d$, governed by some model parameters $\thetab = [\theta_0, \theta_1]^\top$, the offset and the gradient, that we wish to estimate. If we then wish to make $D$ measurements in order to estimate $\thetab$, we construct a design vector $\dbf = [d_1, \dots, d_D]^\top$ that consists of individual experimental designs. For each $d_i$ we have an independent measurement $y_i$, which gives us the data vector $\ybf = [y_1, \dots, y_D]^\top$. There may also be several noise sources affecting the outcome of measurements, complicating the parameter estimation of such a model. If all of these noise sources happen to be Gaussian, we can write down an analytic likelihood and compute the posterior distribution, as well as MI, exactly.

For this toy model, we choose to include two different noise sources, a Gaussian noise source $\mathcal{N}(\epsilon; 0,1)$ and a Gamma noise source $\Gamma(\nu; 2,2)$. The sampling path is given by
\begin{equation} \label{eq:linear_path}
\ybf = \theta_0 \mathbf{1} + \theta_1 \dbf + \bm{\epsilon} + \bm{\nu},
\end{equation}
where $\bm{\epsilon} = [\epsilon_1, \dots, \epsilon_D]^\top$ and $\bm{\nu} = [\nu_1, \dots, \nu_D]^\top$ consist of i.i.d.~samples; $\mathbf{1}$ is a $D$-dimensional vector of ones. While numerical integration could be used to approximate the posterior and MI, we here consider it an implicit model to test our approach. The Jacobian matrix $\Jac$, needed for the computation of the gradient in \eqref{eq:nn_gradd}, is 
\begin{equation} \label{eq:linear_pathgrad}
\Jac = \theta_1 \mathds{1}, 
\end{equation}
where $\mathds{1}$ is the identity matrix. Knowing the sampling path
and its gradient with respect to designs then allows us to compute an
estimate of the gradients $\nabla_\dbf \bound$ in~\eqref{eq:minebed_gradient} and thus find the optimal designs
$\dbf^\ast= [d_1^\ast, \dots, d_D^\ast]^\top$.

At first, we wish to take $1$ measurement only, i.e.~we set $D=1$. We randomly initialise a design $d \in [-10, 10]$ and sample $30{,}000$ parameters from the prior distribution $p(\thetab) = \mathcal{N}(\thetab; \mathbf{0}, 3^2\mathbb{I})$. For every training epoch we generate $30{,}000$ new data samples because we continuously update the design variable and hence the data-generating distribution changes. For the neural network $\NN$ we use one layer of $100$ hidden units, with a ReLU activation layer after the input layer. We update the neural network parameters and designs using the Adam optimiser, with learning rates of $l_{{\psib}}=10^{-4}$ and $l_\dbf=10^{-2}$, respectively. Because the design domain is bounded, we ignore any design updates that would go beyond the specified domain of $d \in [-10, 10]$.

\begin{figure}[!t]
\includegraphics[width=\linewidth]{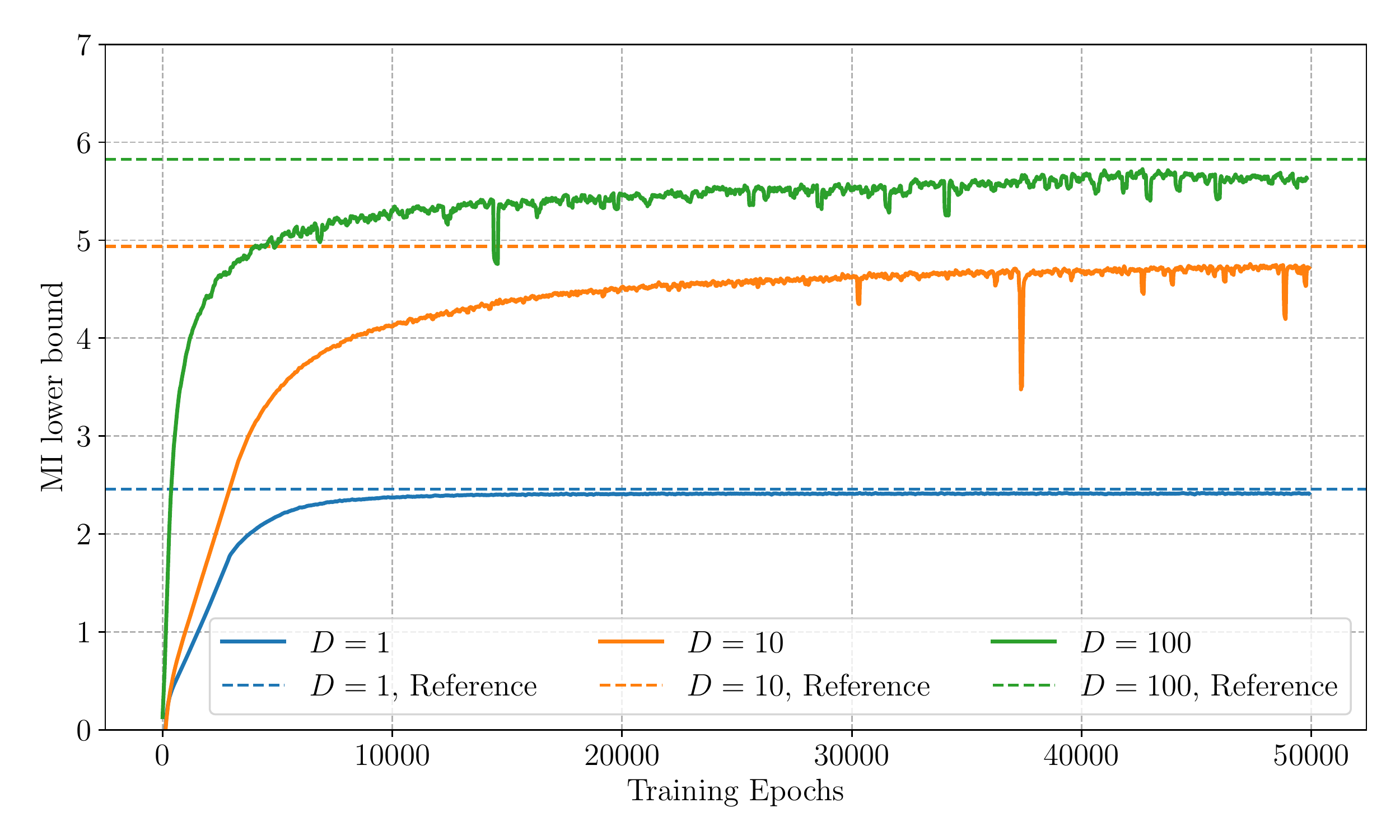}
\caption[]{MI lower bound as a function of neural network training
  epochs for the $D=1$ dimensional (blue), $D=10$ dimensional
  (orange) and $D=100$ dimensional (green) noisy linear model. Shown are the moving averages with a
  window size of $100$ and the dotted lines are numerical reference MI
  values computed at $d^\ast$.}
\label{fig:linear_training_jointdim}
\end{figure}

The blue curve in Figure~\ref{fig:linear_training_jointdim} shows the MI lower bound as a function of training epochs for one-dimensional designs. It can be seen that the lower bound $\bound$ converges after roughly $15{,}000$ epochs. The final MI lower bound is around $2.5$, close to a reference MI computed numerically.\footnote{See the supplementary material for how this is computed.} The optimal design found is $d^\ast = -10$, i.e.~at the boundary; we show the convergence of the design as a function of training epochs in the supplementary material. This optimal design value is intuitive, as there is a larger signal-to-noise ratio at large designs for the linear model.

\begin{figure}[!t]
\includegraphics[width=\linewidth]{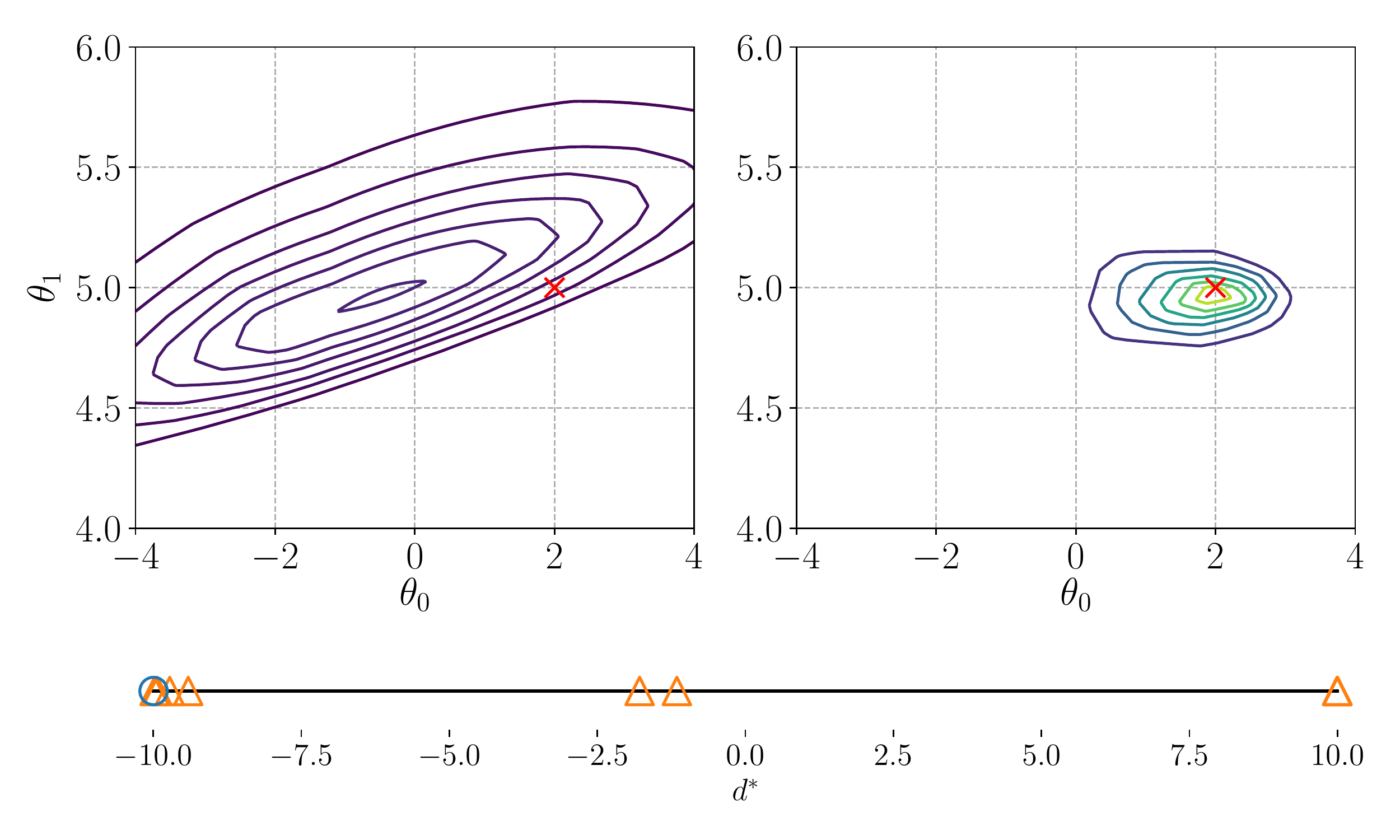}
\caption[]{Posterior density for the one-dimensional (left) and 10-dimensional (right) noisy linear model. Real-world observations $\ybf^\ast$ are generated at $\dbf^\ast$ using $\thetab_{\text{true}} = [2, 5]^\top$ (shown as a red cross). The bottom line shows the optimal designs for $D=1$ (blue circle) and $D=10$ (orange triangles).}
\label{fig:linear_post_jointdim}
\end{figure}

Having found an optimal design $d^\ast$, we can perform a real-world experiment to obtain data $y^\ast$ and use these to estimate the model parameters $\thetab$. We here use $\thetab_{\text{true}} = [2, 5]^\top$ to generate $y^\ast$. Using~\eqref{eq:post_density} we then compute the posterior density, which we show to the left in Figure~\ref{fig:linear_post_jointdim}. The posterior density is relatively broad for both parameters, with the mode being quite far from the ground truth. By means of~\eqref{eq:post_density} and categorical sampling we obtain posterior samples that we use to estimate the parameters, resulting in $\widehat{\theta}_0 = -0.736 \pm 3.074$ and $\widehat{\theta}_1 = 5.049 \pm 0.401$, where the error indicates a 68\% confidence interval. Because the model parameters are two-dimensional, accurately estimating both with one observation is naturally impossible.

We now assume that we have enough budget to take $D=10$ measurements
of the response variable. It is generally worthwhile to do
hyper-parameter optimisation to choose an appropriate neural network
architecture for $\NN$.  By means of grid search (see the
supplementary material), we found a neural network with one hidden
layer and $150$ hidden units to be working best. As before, we used
the Adam optimiser with learning rates of $l_\psi=10^{-4}$ and
$l_{\dbf}=10^{-2}$.  While finding an optimal architecture is usually
difficult, this problem is mitigated by the fact that our method only
requires synthetic data that we can generate ourselves. The resulting
MI lower bound for $D=10$ is shown in
Figure~\ref{fig:linear_training_jointdim} (orange curve), including a
reference MI computed at $\dbf^\ast$. The final MI lower bound is
around $4.5$ with a bias to the reference MI value that is slightly
larger than for $D=1$. We show the optimal designs $\dbf^\ast =
[d_1^\ast, \dots, d_{10}^\ast]^\top$ in the bottom of
Figure~\ref{fig:linear_post_jointdim}. The design dimensions cluster
in three distinct regions: Two clusters are at the two boundaries
which is ideal to learn the slope $\theta_1$, and one cluster is in
the middle where the response is most strongly affected by the offset
$\theta_0$. The supplementary material shows the convergence of each
design dimension as a function of training epochs. As done for the
one-dimensional model, we would then perform a real-world experiment
at $\dbf^\ast$ to observe the response $\ybf^\ast$, which we generate
using $\thetab_{\text{true}} = [2, 5]^\top$, and use this to estimate
the parameters $\thetab$. We show the resulting posterior density,
again computed by using~\eqref{eq:post_density}, to the right in
Figure~\ref{fig:linear_post_jointdim}. Due to having more
measurements, the posterior distribution is narrower and more accurate
for $D=10$ than for $D=1$. By using categorical sampling we obtain
the estimates $\widehat{\theta}_0 = 1.441 \pm 0.826$ and
$\widehat{\theta}_1 = 4.949 \pm 0.108$.

In order to test the scalability and robustness of our method, we
further consider the noisy linear model for $D=100$ dimensions. We
show the corresponding mutual information lower bound in
Figure~\ref{fig:linear_training_jointdim} as the green curve. The
final mutual information lower bound for $D=100$ is higher than for
the $D=10$ or $D=1$ models, which should be expected due to having
more measurements. Even in such high-dimensions, we find that we can
converge to a lower bound that only has a small bias with respect to a
reference mutual information value (shown as a dashed, green
line). Interestingly, we find that extrapolating the design strategy
from $D=10$ to $D=100$ yields sub-optimal results. Instead, one should
focus measurements around zero and avoid the domain boundaries. See
the supplementary materials for the results, the description of the
setup, and further analysis and figures.

\subsection{Pharmacokinetic Model} 

In this experiment we are concerned with finding the optimal blood sampling time during a pharmacokinetic (PK) study. These types of studies involve the administration of a drug to a patient or group of patients and then measuring the drug concentration at certain times to study the underlying kinetics. We shall use the compartmental model introduced by~\citet{Ryan2014} to simulate the drug concentration at a particular time. Their model is governed by three parameters that we wish to estimate: the absorption rate $k_a$, the elimination rate $k_e$, and the volume of distribution $V$ which indicates how the drug will distribute in the body. We will assume that we have a group of $D$ patients but for each of the patients, we can take only one blood sample at time $t$.

The model consist of a latent variable $z(t)$ for the drug concentration at design time $t$,
\begin{align} \label{eq:pharma_latent}
z(t) = \frac{D_V}{V} \frac{k_a}{k_a - k_e} \left[ e^{-k_et} - e^{-k_at} \right](1+\epsilon),
\end{align}
where $D_V=400$ is a constant and $\epsilon \sim \mathcal{N}(0,0.01)$ is multiplicative noise modelling heteroscedasticity that is often observed in drug concentration data. We further have an additive observation model that yields the measured drug concentration $y$ at design time $t$, i.e.
\begin{align} \label{eq:pharma}
y(t) = z(t) + \nu,
\end{align}
where $\nu \sim \mathcal{N}(0,0.1)$. We note that \eqref{eq:pharma_latent} is a simplified model in that it assumes that the parameters $\thetab = [k_a, k_e, V]^\top$ are the same for the group of patients considered, and that it does not model the time-course of the concentration for each patient individually but only the marginal distribution at design time $t$. If we wanted to take several measurements per patient instead of one, we had to model the concentration correlation across time and e.g.\ use models described by stochastic differential equations \citep[see e.g.\ ][]{Donnet2013}.

We follow \citet{Ryan2014} and assume the following prior for $\thetab$
\begin{equation} \label{eq:prior}
\log \thetab \sim \mathcal{N}   
\left( 
	\begin{bmatrix}
		\log 1 \\
		\log 0.1 \\
		\log 20 \\
    \end{bmatrix}
,
\begin{bmatrix}
	0.05 & 0 & 0 \\
	0 & 0.05 & 0 \\
	0 & 0 & 0.05 \\
\end{bmatrix}
\right),
\end{equation}
with the additional constraint that $k_a > k_e$, enforced via rejection sampling. The PK model in~\eqref{eq:pharma} can be written in terms of the sampling path $y = h(\epsilon, \nu \mid \thetab, t)$. The corresponding sampling path gradients are
\begin{align}
\frac{\partial y(t)}{\partial t} &= \frac{D_V}{V} \frac{k_a}{k_a - k_e} \left[-k_e e^{-k_et} + k_a e^{-k_at} \right](1+\epsilon) \label{eq:pharma_grads}
\end{align}
Knowing the sampling path derivative in~\eqref{eq:pharma_grads} allows us to find optimal blood sampling times by gradient ascent on the MI lower bound. Similar to the linear model, we here assume that we wish to take $D$ measurements. In practice, this means that we have $D$ patients and take one blood sample per patient. Our design vector then becomes $\dbf=[t_1, \dots, t_D]^\top$, with corresponding observations $\ybf=[y_1, \dots, y_D]^\top$.

We start with the simpler setting of only performing one experiment, i.e.~we set $D=1$. At first, we randomly initialise a design time in $t \in [0, 24]$ hours, the design domain for this experiment. We then draw $30{,}000$ model parameter samples $\thetab^{(i)}$ from the prior distribution in~\eqref{eq:prior}. At every epoch we simulate corresponding data samples $y^{(i)}$ using the current design $t$ and the sampling path in~\eqref{eq:pharma}. For $\NN$ we use a neural network with one hidden layer of $100$ hidden units, as well as a ReLU activation function after the input layer. For the simpler setting of one-dimensional designs, we do not perform hyper-parameter optimisation of the network architecture. We then follow the procedure outlined in Section~\ref{sec:minebed} to train the neural network in order to obtain an optimal design time. We use the Adam optimiser to learn ${{\psib}}$ and $t$ with learning rates of $l_{{\psib}} =10^{-3}$ and $l_t = 10^{-2}$, respectively.

\begin{figure}[!t]
\includegraphics[width=\linewidth]{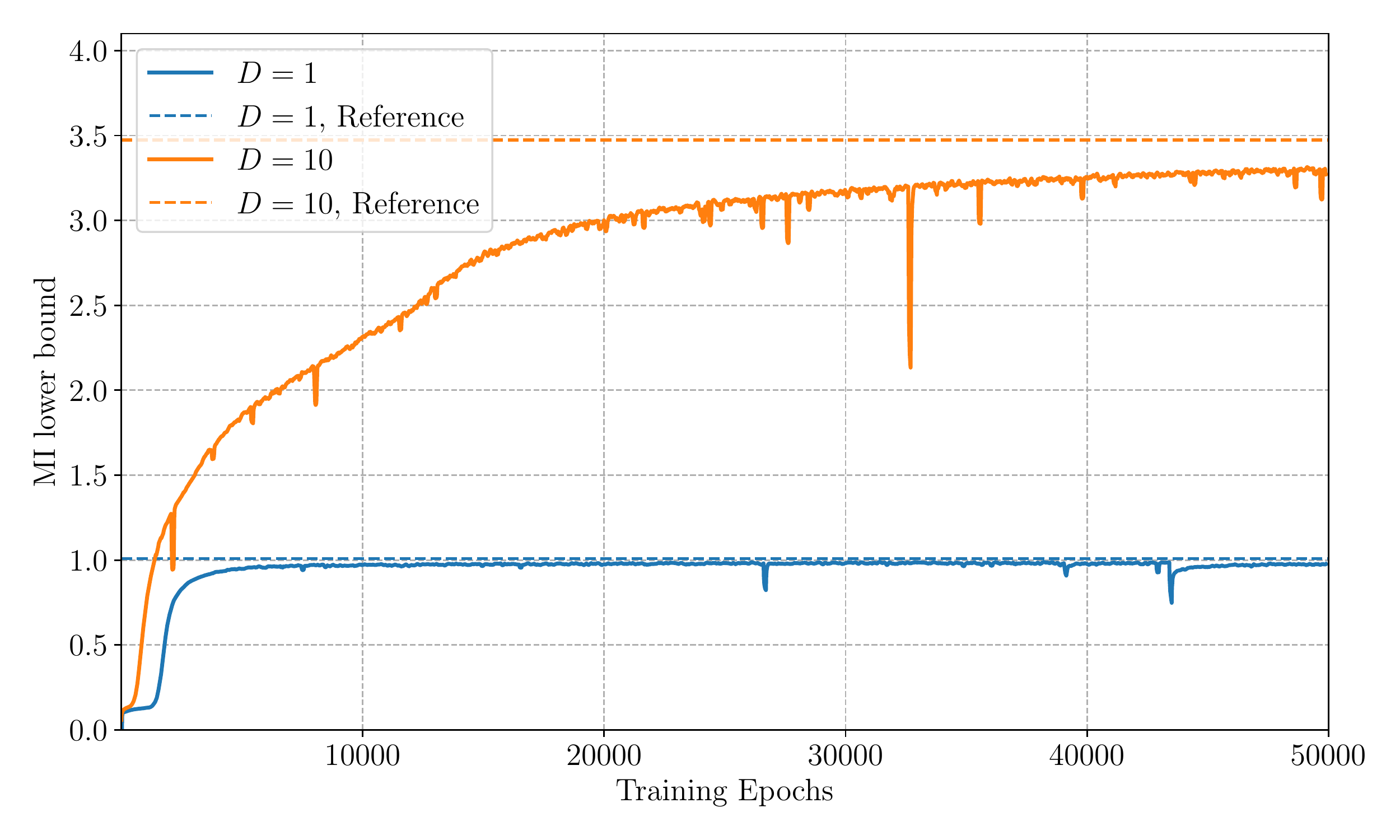}
\caption[]{MI lower bound as a function of neural network training epochs for the one-dimensional (blue) and 10-dimensional (orange) PK model. Shown are the moving averages with a window size of $100$ and the dotted lines are numerical reference MI values computed at $t^\ast$.}
\label{fig:pharma_training}
\end{figure}

In Figure~\ref{fig:pharma_training} we show the MI lower bound as a
function of training epochs for the simple one-dimensional PK model
(blue curve). The MI lower bound converges to a value of about $1$ and
has a negligibly small bias compared to a reference MI value
at the optimal design $t^\ast = 0.551$. In the supplementary material,
we give details on the computation of the reference value and
illustrate the convergence of the design as a function of training
epochs.

\begin{figure}[!t]
\includegraphics[width=\linewidth]{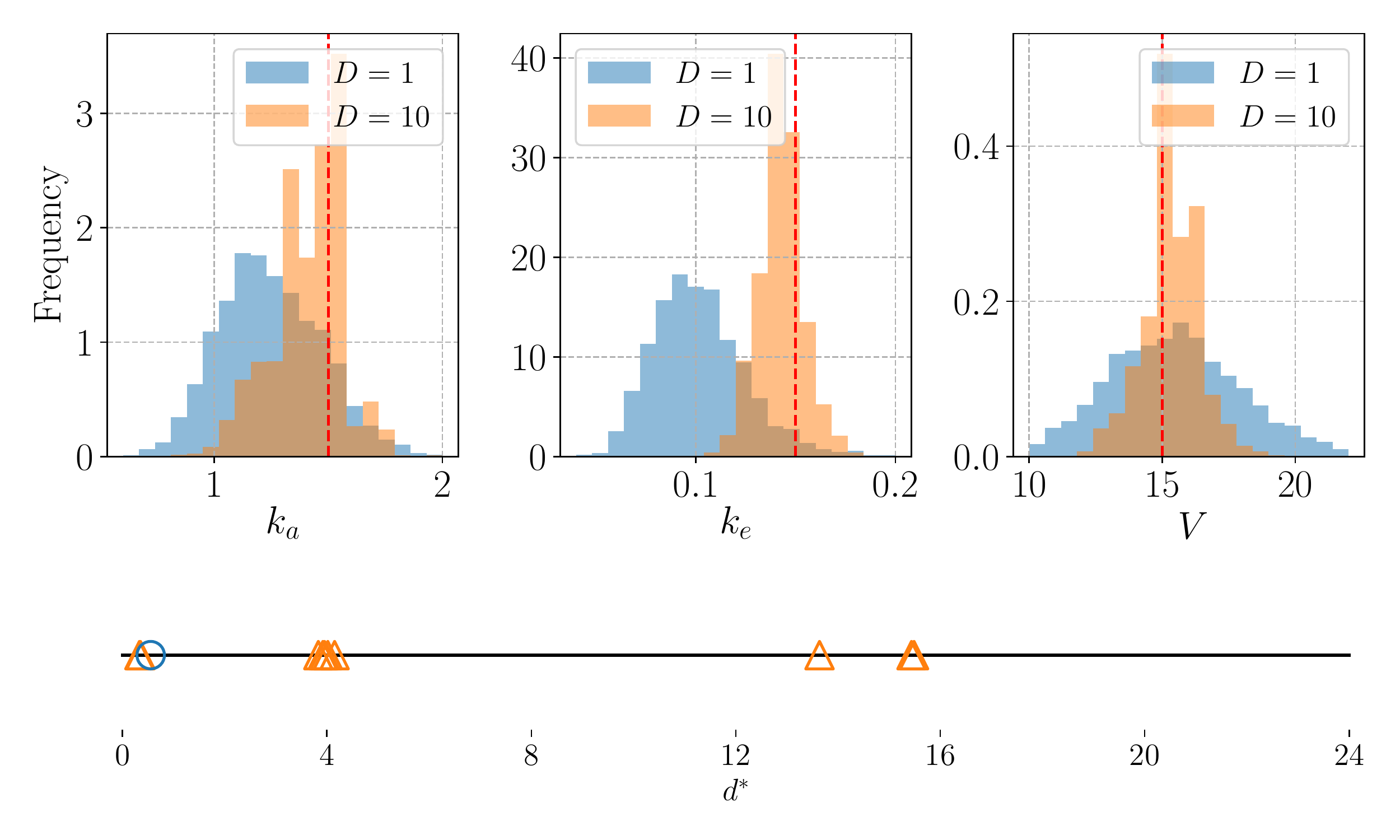}
\caption[]{Posterior marginal distributions of the model parameters for the one-dimensional (blue) and 10-dimensional (orange) PK model, with red-dotted lines showing the true model parameters. The bottom line shows the optimal designs for $D=1$ (blue circle) and $D=10$ (orange triangles).}
\label{fig:pharma_post}
\end{figure}

Using the optimal design $t^\ast$ we would then do a real-world experiment to obtain data $y^\ast$. Here, we generate $y^\ast$ using $\theta_{\text{true}} = [1.5, 0.15, 15]^\top$. Using the trained neural network $T_{\psib^\ast}(\thetab, \ybf)$, as well as~\eqref{eq:post_density}, we then compute the posterior density $p(\thetab \mid t^\ast, y^\ast)$ and use categorical sampling to obtain posterior samples. The marginal posterior distribution of each model parameter is shown in Figure~\ref{fig:pharma_post}. The posterior mean is close to its true value for $V$, less close for $k_a$, but relatively far for $k_e$. This is because with only one measurement at low design times we can only quite accurately determine the ratio of $k_a/V$, which can be seen from their joint distribution shown in the supplementary material.

We now consider a more complex setting of performing $D=10$
measurements. As before, appropriate choices of neural network
architectures are not obvious. Thus, we test one-layer neural networks
with different number of hidden units, as well as different learning
rate schedules. We found that the best-performing neural network had a
hidden layer of size $300$ and a multiplicative learning rate schedule
with a multiplying factor of $0.8$. In practice, this means that we
multiply the initial learning rates $l_{\psib}=10^{-3}$ and
$l_{\dbf}=10^{-2}$ by $0.8$ every $5{,}000$ epochs. We again note that
further hyper-parameter search would be possible. Similar to before we
use $30{,}000$ prior samples, and corresponding new data generated at
each epoch, to train the neural network. We show the resulting lower
bound $\bound$ as a function of training epochs in
Figure~\ref{fig:pharma_training} (orange curve). The neural network
for $D=10$ takes longer to converge than for $D=1$ and ends up at a
higher MI lower bound; this is intuitive as more data naturally yields
more information about the model parameters. The bias of the final MI
lower bound to a reference MI value is larger for $D=10$ than for
$D=1$ due to the increased dimensions but it is still reasonably
small. The optimal designs, shown in the bottom of
Figure~\ref{fig:pharma_post}, form three clusters similar to what has
been observed by \citet{Ryan2014}. The early cluster allows us to
determine the ratio $k_a/V$ as before, the late cluster reduces the
effect of $k_a$, while the middle cluster, which is needed to identify
the third parameter, corresponds to where the mean response tends to
be highest \citep{Ryan2014}. The convergence of the designs are shown
in the supplementary material. Similarly to before, we then take a
real-world measurement $\ybf^\ast$ at $\dbf^\ast$, simulated using
$\theta_{\text{true}} = [1.5, 0.15,
  15]^\top$. Using~\eqref{eq:post_density} and categorical sampling we
obtain samples from the posterior distribution, with resulting
marginal samples for each model parameter shown in
Figure~\ref{fig:pharma_post}.\footnote{See the supplementary material
  for the joint distributions.} As opposed to $D=1$, where we could
only estimate the ratio $k_a/V$ well, for $D=10$ we can estimate all
parameters more precisely and accurately.

\subsection{Locating a Gas Leak}

In this experiment we aim to locate a gas leak in a two-dimensional space. In order to find the source $[\theta_x, \theta_y]^\top$ of the leak, we need to decide at which optimal position $\mathbf{d} = [d_x, d_y]^\top$ to measure the gas concentration $g$. We here use the fluid simulator of~\citet{Asenov2019} to forward-simulate the movement of gas particles. This fluid simulator solves the Navier-Stokes equations to yield gas concentrations $g$ over a two-dimensional grid as a function of time $t$. Throughout this experiment we shall be assuming a steady-state solution, which is achieved by ignoring early time-steps. By using the translational properties of their simulator we can compute $g$ at any position $\mathbf{d}$ for any source location $[\theta_x, \theta_y]^\top$ using only one forward simulation~\citep[see][for more information]{Asenov2019}, greatly speeding up computation times. We assume that our measurements are subject to additive Gamma observational noise $\eta \sim \Gamma(2,0.2)$. In the fluid simulator, we choose to set the gas viscosity to $0$, the gas diffusion to $10^{-4}$, and set the wind speed to $0.1$. Furthermore, the discrete grid size is set to $64 \times 64$; a finer grid would yield more accurate localisations, at the cost of higher computation times.

While we keep most parameters of the fluid simulator constant, we assume uncertainty over the wind direction $W_d$, which introduces additional stochasticity. To do so, we put a (discretised) Gaussian prior over the wind direction $W_d$, centered at $45^\circ$ from east with a standard deviation of $5^\circ$. In practice, this means that we define the wind direction over a grid $W_d \in \{30.0, 32.5, \dots, 60.0\}$ with a corresponding set of prior probabilities given by $p(W_d) \propto \mathcal{N}(W_d; 45, 5^2)$. Including this nuisance parameter, the model parameters for this problem are now given by $\thetab = [\theta_x, \theta_y, W_d]^\top$.

\begin{figure}[!t]
\includegraphics[width=\linewidth]{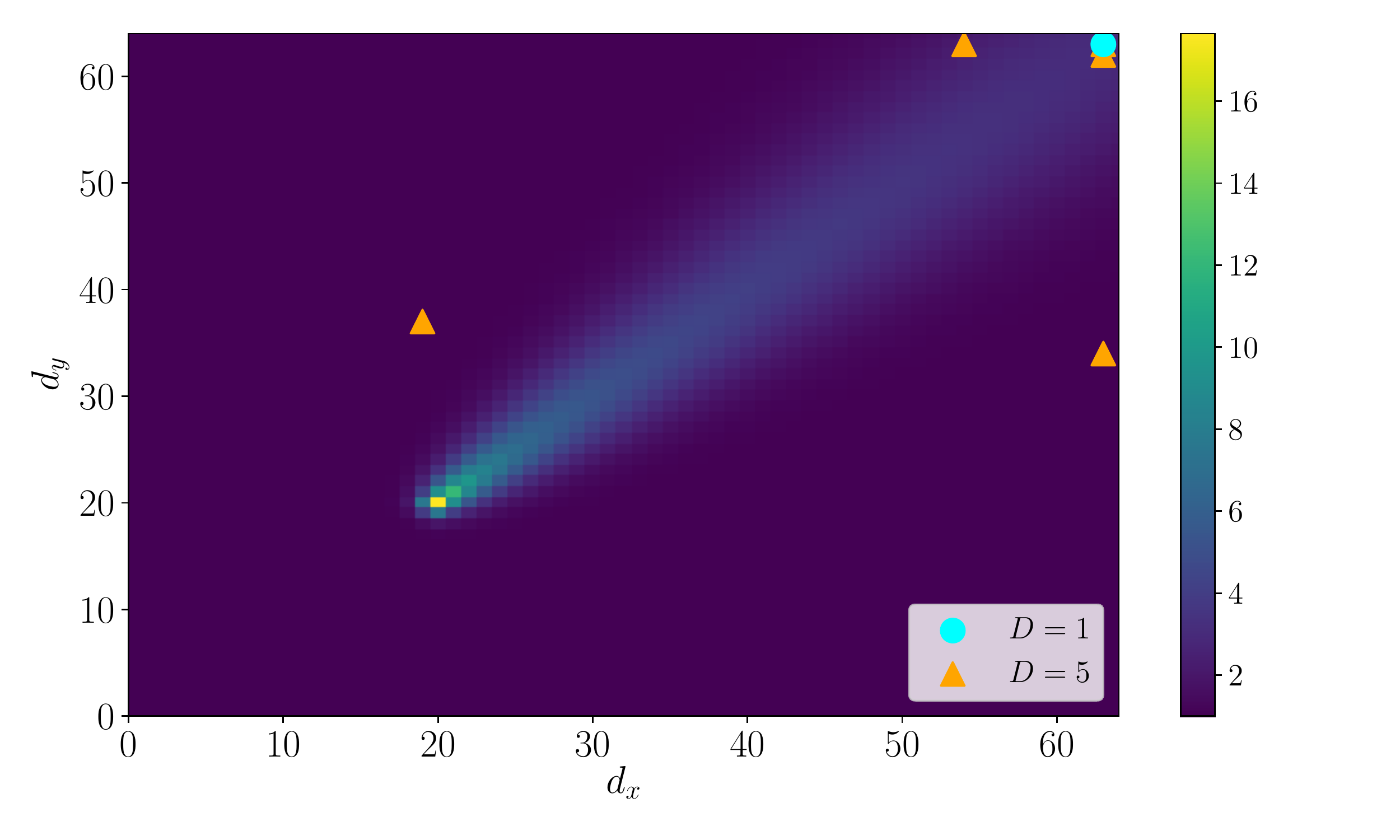}
\caption[]{Marginal gas concentration as a function of $\dbf$ with the gas leak at the true source location. Also shown are the optimal designs for $D=1$ measurement (cyan circles) and $D=5$ measurements (orange triangles).}
\label{fig:gasleak_opt}
\end{figure}

For this particular fluid simulator, we do not have gradients of the
sampling path. Thus, we have to revert to using Bayesian optimisation
to optimise the design variable $\mathbf{d}$, as explained in
Section~\ref{sec:bo}. In order to train $\NN$ at a particular design
$\dbf$ we use a one-layer neural network with $100$ hidden units and a
ReLU activation function after the input layer. The neural network
parameters ${{\psib}}$ are still optimised by means of the Adam
optimiser with a learning rate of $l_{{\psib}} = 10^{-3}$. Before
training, we simulate $10{,}000$ uniform prior samples
$\thetab^{(i)}$. During every re-training of $\NN$ at a different
design $\mathbf{d}$ we then simulate $10{,}000$ corresponding samples
of the gas concentration $g$. This procedure is repeated until we
converge to the optimal design of $\dbf^\ast = [64, 64]^\top$. We
visualise the optimal design location in Figure~\ref{fig:gasleak_opt}
and the supplementary material shows a plot of the GP posterior mean
of the MI lower bound after training. The optimal design matches the
fact that the wind is pointing north east, hence making upstream
designs in that direction more informative. To illustrate this, in
Figure~\ref{fig:gasleak_opt} we also show the gas concentration on a
grid of $\dbf$, marginalised over the wind speed, using $[\theta_x,
  \theta_y]^\top_{\text{true}} = [20, 20]^\top$ as the true source
location.

\begin{figure}[!t]
\includegraphics[width=\linewidth]{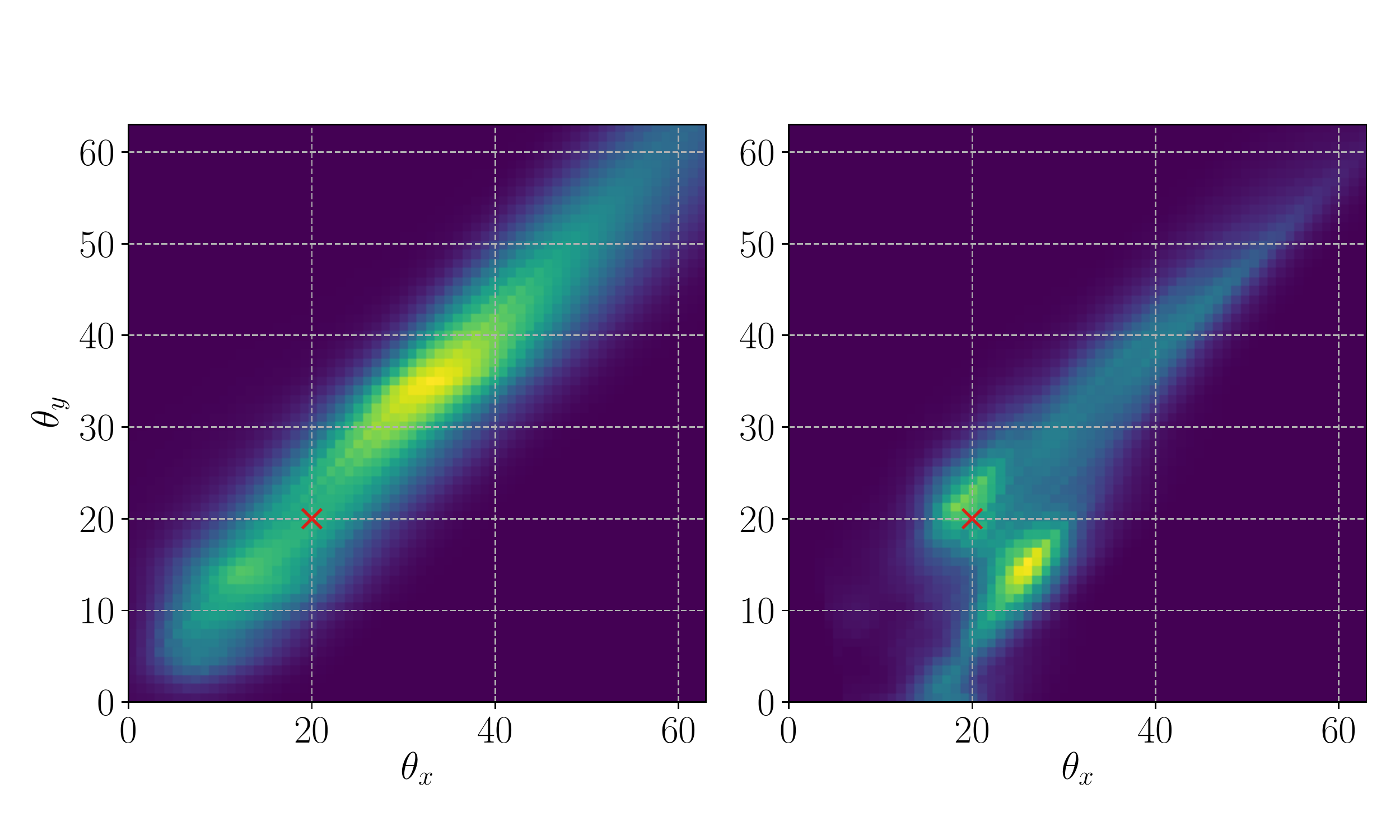}
\caption[]{Marginal Posterior density for the source locations of the gas leak with one measurement (left) and five measurements (right). Shown as the red cross is the true gas leak location.}
\label{fig:gasleak_post}
\end{figure}

Similar to before, given an optimal design $\mathbf{d}^\ast$ we can make a real-world measurement of the gas concentration $g^\ast$, generated using $[\theta_x, \theta_y]^\top_{\text{true}} = [20, 20]^\top$ and a true wind direction of $45^\circ$. We then use the trained network $T_{{{\psib}}^\ast}(\thetab, g)$ at the optimal design $\mathbf{d}^\ast$ to obtain $p(\thetab \mid g^\ast, \dbf^\ast)$ without any additional computation using~\eqref{eq:post_density}. We evaluate this density over the $64\times64$ grid of possible gas leak locations, as shown in the left in Figure~\ref{fig:gasleak_post}, where we have marginalised out the wind direction because it is a nuisance variable. While broad, the posterior accurately captures the direction of the gas leak relative to the measurement location.

As can be seen in the left in Figure~\ref{fig:gasleak_post}, accurately locating a gas leak with only one measurement is extremely difficult. We wish to rectify this by increasing the number of measurements to $D=5$. As a consequence, our design variable $\mathbf{d} = [d_{1,x}, d_{1,y}, \dots, d_{5, x}, d_{5, y}]^\top$ is now $10$-dimensional, with corresponding gas concentration measurements of $\mathbf{g} = [g_1, \dots, g_5]^\top$. As before, we test different architectures and find out that a one-layer neural network with $200$ hidden units has the best performance; all other settings are kept as before. Repeating the BO optimisation procedure we obtain an optimal design $\mathbf{d}^\ast$ that we visualise in Figure~\ref{fig:gasleak_opt}. The optimal designs are spread out at the edges of the grid where the wind is pointing to. This is again intuitive, as we want to be measuring upstream to ensure that we do not get concentration measurements of zero. Furthermore, we have one optimal design away from the edges that helps us localise the gas leak better. Depending on the starting position of the experimenter, one could then determine the optimal flight path to take all of these measurements.

Taking real-world observations at $\dbf^\ast$ yields gas concentration
measurements $\mathbf{g}^\ast$. Using these and the trained neural
network we compute the posterior density by means
of~\eqref{eq:post_density}, which we show in the right in
Figure~\ref{fig:gasleak_post}. The posterior distribution is more
concentrated for $5$ measurements than for $1$ measurement. While it is bimodal, the true source
location is covered by one of the modes. This bimodality arises
because of the uncertainty in the wind direction; if we know the wind
direction, the posterior is unimodal (see the supplementary material).

\section{Conclusions}

In this work we have presented a novel approach to Bayesian experimental design for implicit models, where the data-generation distribution is intractable but sampling is possible. Our method uses a neural network to maximise a lower bound on the mutual information between data and parameters. This allows us to find both the optimal experimental designs and the corresponding posterior distribution by training a single neural network.

Whenever we have access to the gradients of the data sampling path of
the implicit model, we can update neural network parameters and
designs at the same time using gradient ascent, greatly enhancing
scalability. While this is possible for a large and interesting class
of implicit models, in situations where gradients with respect to the
designs are not available, we can still update the neural network
parameters using gradient ascent and the designs by means of e.g.\ Bayesian
Optimisation.

In a variety of simulation studies, we found that our method provides suitable and intuitive experimental designs, as well as reasonable posterior distributions, even in higher design dimensions.

It would be interesting to apply our method to sequential experimental design where we can update the posterior of the model parameters after each real-world experiment performed. 
While there is some work on extending sequential experimental design to the significantly more difficult area of implicit models~\citep[e.g.][]{Kleinegesse2020, Hainy2016}, we believe that there is still much to explore, especially in the context of our method.
Additionally, it might be worthwhile to investigate the extension of our method to the setting of experimental design for model discrimination or the prediction of future observations.


\section*{Acknowledgements}

Steven Kleinegesse was supported in part by the EPSRC Centre for Doctoral Training in Data Science, funded by the UK Engineering and Physical Sciences Research Council (grant EP/L016427/1) and the University of Edinburgh. 





\bibliography{references}
\bibliographystyle{icml2020}

\cleardoublepage
\appendix
\twocolumn 

\twocolumn[
\icmltitle{Supplementary Material for \\ Bayesian Experimental Design for Implicit Models \\ by Mutual Information Neural Estimation}
]

\section{Additional Information for the Linear Model}

\subsection{Reference MI Calculation} \label{sec:ref}

In order to compute a reference mutual information (MI) value at
$\dbf^\ast$, we rely on a nested Monte-Carlo sample average of the MI
and an approximation to the likelihood
$p(\ybf \mid \dbf^\ast, \thetab)$.

In the setting where we wish to make $D$ independent measurements at $\dbf^\ast = [d_1^\ast, \dots, d_D^\ast]^\top$, the likelihood factorises as $p(\ybf \mid \dbf^\ast, \thetab) = \prod_{j=1}^D p(y_j \mid d_j^\ast, \thetab)$. Using this and by means of a sample-average of the marginal $p(\ybf \mid \dbf^\ast)$, we can approximate the MI (as shown in the main text) as follows,
\begin{equation} \label{eq:mi_approx}
I(\dbf^\ast) \approx \frac{1}{N} \sum_{i=1}^N \left[ \log{\frac{\prod_{j=1}^D p(y_j^{(i)} \mid d_j^\ast, \thetab^{(i)})}{\frac{1}{M}\sum_{s=1}^M\prod_{j=1}^D p(y_j^{(i)} \mid d_j^\ast, \thetab^{(s)})}} \right],
\end{equation}
where $y_j^{(i)} \sim p(y_j \mid d_j^\ast, \thetab^{(i)})$, $\thetab^{(i)} \sim p(\thetab)$ and $\thetab^{(s)} \sim p(\thetab)$.

In order to be able to compute the MI approximation
in~\eqref{eq:mi_approx} for the linear model, we need to built an
approximation to the density $p(y_j \mid d_j^\ast, \thetab)$. The
sampling path for the linear model is given by $y_j = \theta_0
+ \theta_1d^\ast_j + \epsilon + \nu$, where
$\epsilon \sim \mathcal{N}(0,1)$ and $\nu \sim \Gamma(2,2)$ are
sources of noise. The distribution $p_{\text{noise}}$ of $\epsilon
+ \nu$ is given by the convolution of the densities of $\epsilon$ and
$\nu$. It could be computed via numerical integration. Here we compute
it by a Kernel Density Estimate (KDE) based on $50{,}000$ samples of
$\epsilon$ and $\nu$. By rearranging the sampling path to $y_j -
(\theta_0 + \theta_1d) = \epsilon + \nu$ we then obtain that
$p(y_j \mid d_j, \thetab) = p_{\text{noise}}(y_j - (\theta_0
+ \theta_1d))$, allowing us to estimate the MI
using~\eqref{eq:mi_approx}. We here use $N=5{,}000$ and $M=500$.

\subsection{Hyper-Parameter Optimisation}

\begin{figure}[!t]
\includegraphics[width=\linewidth]{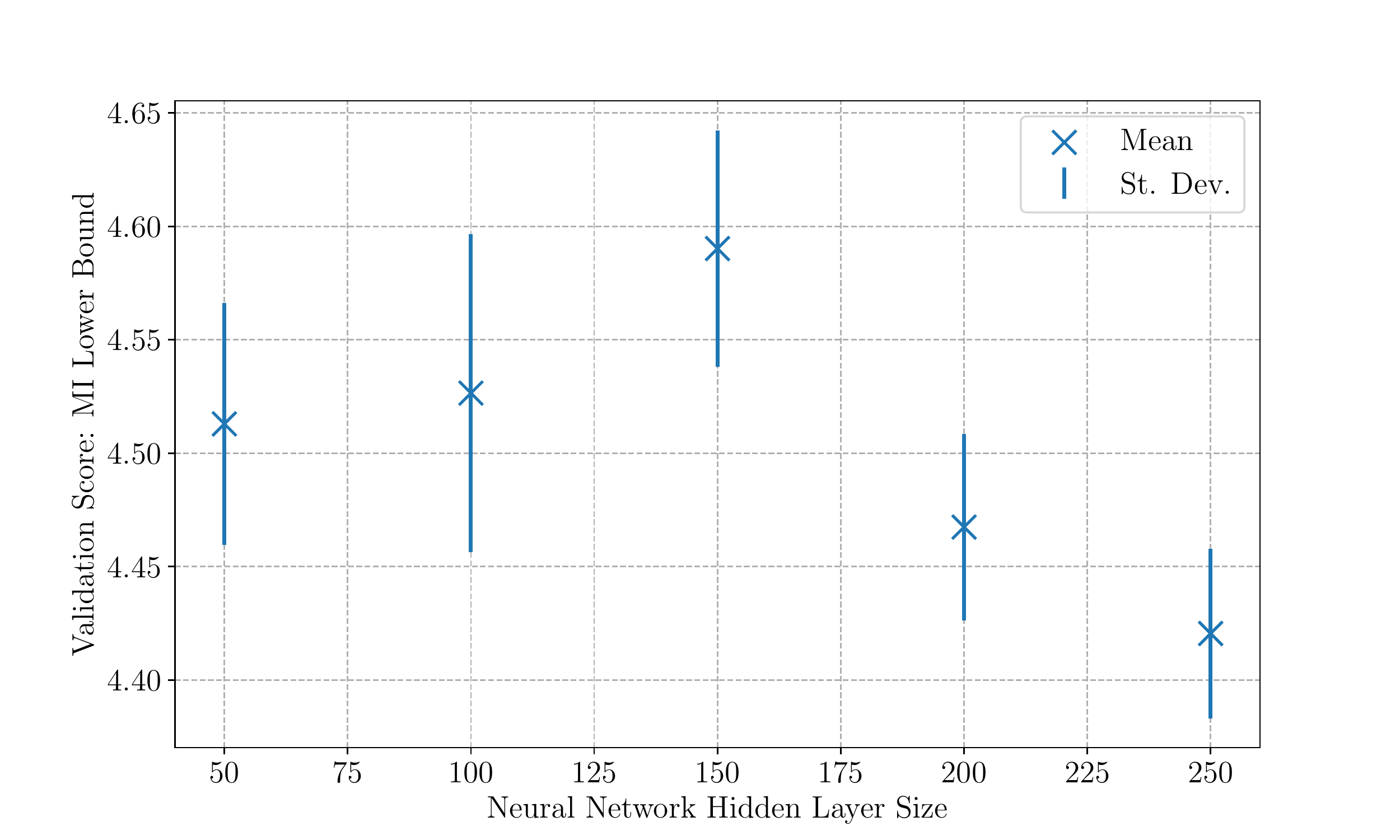}
\caption[]{Mean and standard deviation of validation scores (MI lower bound) for the 10-dimensional noisy linear model, using one-layer neural networks with different number of hidden units.}
\label{fig:linear_D10_hyper}
\end{figure}

We wish to find the optimal neural network size for the 10-dimensional
linear model. To do so, we train several neural networks with one
hidden layer of sizes $H \in \{50,100,150,200,250\}$ for $50{,}000$
epochs using learning rates $l_\psi = 10^{-4}$ and $l_d=10^{-2}$. By
generating samples from the prior distribution and the data-generating
distribution at the optimal design, we can build a validation set that
we use to obtain an estimate of the MI lower bound. Repeating this
several times for every trained neural network yields mean MI lower
bound estimates, as well as standard deviations, as shown in
Figure~\ref{fig:linear_D10_hyper}. Given that we wish to obtain the
maximum MI lower bound, we see that a neural network of size $H=150$ is most
appropriate in our setting.

\subsection{Convergence of Designs}

\begin{figure}[!t]
\includegraphics[width=\linewidth]{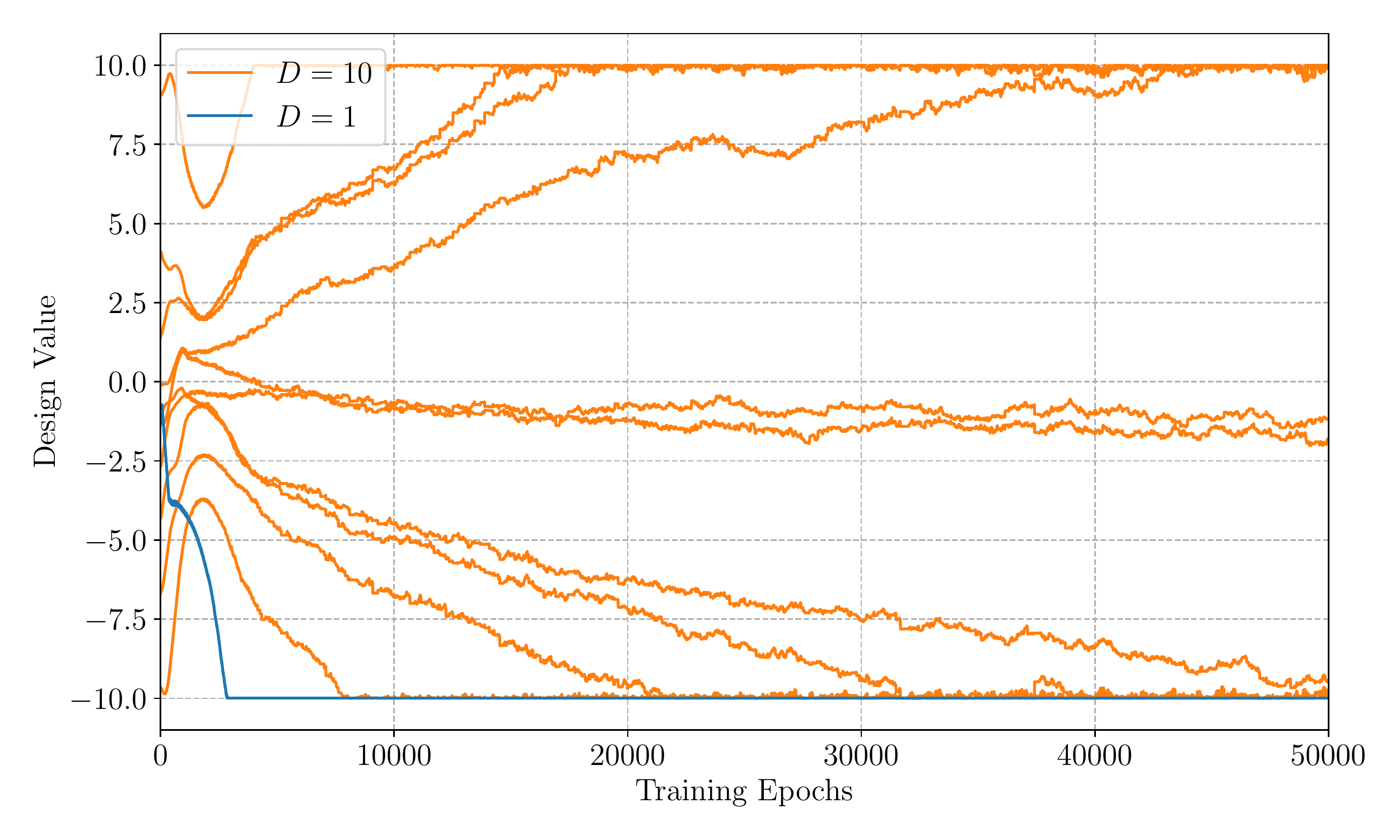}
\caption[]{Convergence of the individual design dimensions for the one-dimensional (blue curve) and 10-dimensional (orange curves) linear model. Note how for the 10-dimensional linear model the design dimensions end up in three different clusters.}
\label{fig:linear_convergence}
\end{figure}

In Figure~\ref{fig:linear_convergence} we show the convergence of the individual design dimensions for the one-dimensional and 10-dimensional linear model. The one-dimensional design converges quickly to the optimal design, while the different dimensions of the 10-dimensional design converge more slowly and end up in three clusters. Two clusters of optimal designs are at the boundaries where the signal-to-noise ratio is highest, allowing us to estimate the slope of the linear model well. The other cluster is near zero, reducing the effect of the slope and allowing us to estimate the offset better.

\subsection{100-Dimensional Linear Model}

In order to test the scalability of our method, we here apply MINEBED to a 100-dimensional version of the linear model. Because of the higher dimensionality of both the data vector $\mathbf{y}$ and the design vector $\mathbf{d}$, we require a neural network with more parameters to obtain a tight bound. We found that a deep neural network, i.e.~more layers with less hidden units, seemed to work better than a wide neural network, i.e.~less layers with more hidden units. Hence, we opted to use a 5-layered network with 50 hidden units for each layer. We train the network with $10{,}000$ samples and use the Adam optimiser, with initial learning rates of $l_{\psi}=10^{-4}$ and $l_{\mathbf{d}}=10^{-2}$.

\begin{figure}[!t]
\includegraphics[width=\linewidth]{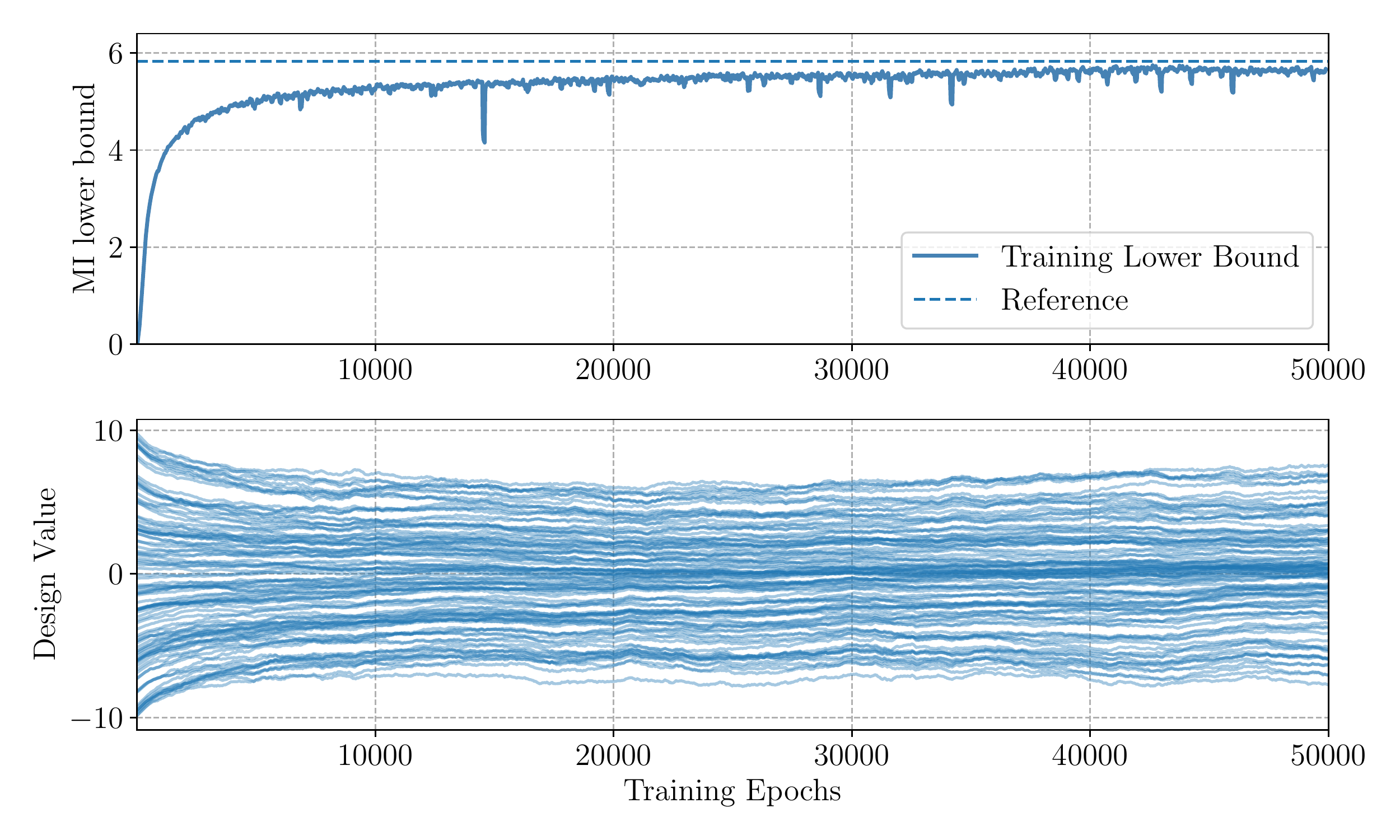}
\caption[]{Convergence of the mutual information lower bound (top) and individual design dimensions (bottom) for the 100-dimensional linear model. The dashed line shows a reference mutual information value at the final, optimal design.}
\label{fig:linear_d100}
\end{figure}

The convergence of the mutual information lower bound and the dimensions of the design vector are shown in the top and bottom of Figure~\ref{fig:linear_d100}, respectively. Also shown is a reference mutual information value computed as explained in Section~\ref{sec:ref}. The mutual information lower bound converges smoothly to a value that is higher than for the 1-dimensional and 10-dimensional version of the linear model (see the main text). This is intuitive, as more data allows us to gain more information about the model parameters. The final lower bound is relatively tight, as it is close to the reference MI value. As can be seen from the bottom of Figure~\ref{fig:linear_d100}, the different dimensions of the design vector converge to the region $d_j \in [-8, 8]$, with more designs centred around zero. Interestingly, this is a different strategy as for 1 and 10 design dimensions, see Figure~\ref{fig:linear_convergence}, where most optimal designs were found to be at the domain boundaries $d_j = -10$ and $d_j = 10$. 

In order to ascertain whether or not this experimental design strategy is sensible, we compute reference MI values for different strategies: 1) designs only clustered at the boundaries ($\text{MI}=4.61$), 2) designs clustered at the boundaries and at zero ($\text{MI}=4.83$), akin to the optimal design for 10 dims, 3) equally spaced designs ($\text{MI}=4.91$) and 4) random designs (average of $4.92$, standard deviation of $0.08$ and maximum value of $5.11$ for $100$ repeats). These values are smaller than the value we obtained ($\text{MI}=5.78$), indicating that we may have found a (locally) optimal design. This further implies that extrapolating the design strategy from the 10-dimensional to the 100-dimensional linear model is sub-optimal. If the experimenter knows that they can make that many measurements, centring them around zero with some dispersion allows for better parameter estimation.

\begin{figure}[!t]
\includegraphics[width=\linewidth]{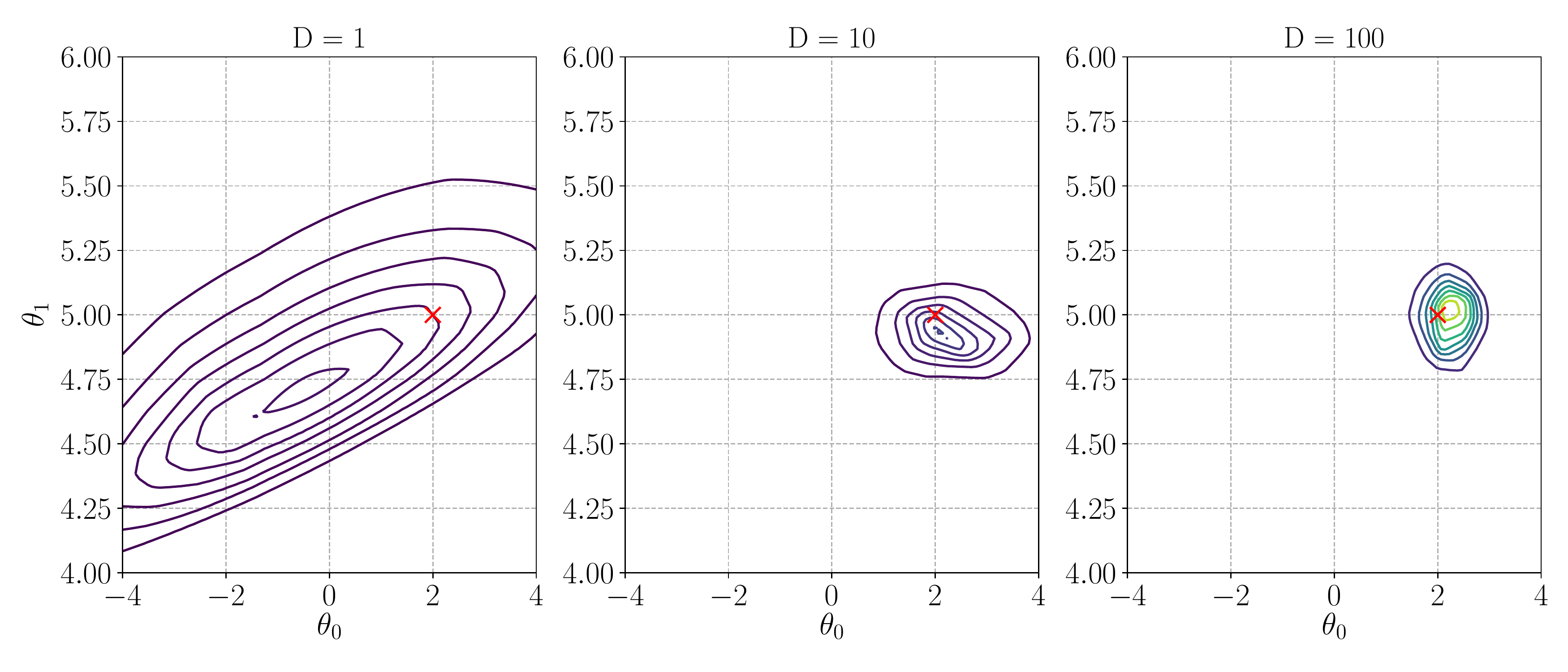}
\caption[]{Comparison of posterior densities obtained for the 1-dimensional (left), 10-dimensional (middle) and 100-dimensional (right) noisy linear model. The red cross shows the true model parameters.}
\label{fig:linear_d100_params}
\end{figure}

In Figure~\ref{fig:linear_d100_params} we show a comparison of the posterior distributions obtained for the 1-dimensional (left), 10-dimensional (middle) and 100-dimensional (right) noisy linear model. The posterior densities were computed using the trained neural network, as explained in the main text. We find that the posterior distribution becomes narrower, with modes that are closer to the true model parameters, as we increase the design dimensions. This is again intuitive, as more measurements allow us to estimate the model parameters better. Overall, these results show that we can effectively find optimal designs and compute corresponding posterior densities even for 100-dimensional experimental design problems.

\section{Additional Information for the PK Model}

\subsection{Reference MI Calculation}

Just like for the linear model, we use the nested Monte-Carlo
approximation of MI given in~\eqref{eq:mi_approx} to compute a
reference MI at $\dbf^\ast = [t_1^\ast, \dots, t_D^\ast]^\top$ for the
pharmacokinetic (PK) model. Even though computing the MI is
intractable, we can write down an equation for the data-generating
distribution $p(y_j \mid t_j, \thetab)$, where $\thetab=[k_a, k_e,
V]^\top$, as both noise sources are Gaussian,
\begin{equation}
p(y_j \mid t_j, \thetab) = \mathcal{N}(y_j; f(t_j, \thetab), f(t_j, \thetab)^2 0.01^2 + 0.1^2),
\end{equation}
where the function $f(t_j, \thetab)$ is given by
\begin{equation}
f(t_j, \thetab) = \frac{D_V}{V} \frac{k_a}{k_a - k_e} \left[ e^{-k_et_j} - e^{-k_at_j} \right].
\end{equation}
Using this expression and a sample average of $p(\ybf \mid \dbf^\ast)$ we can then compute a numerical approximation of the MI given in~\eqref{eq:mi_approx}. We here use $N=5{,}000$ and $M=500$.

\subsection{Hyper-Parameter Optimisation}

\begin{figure}[!t]
\includegraphics[width=\linewidth]{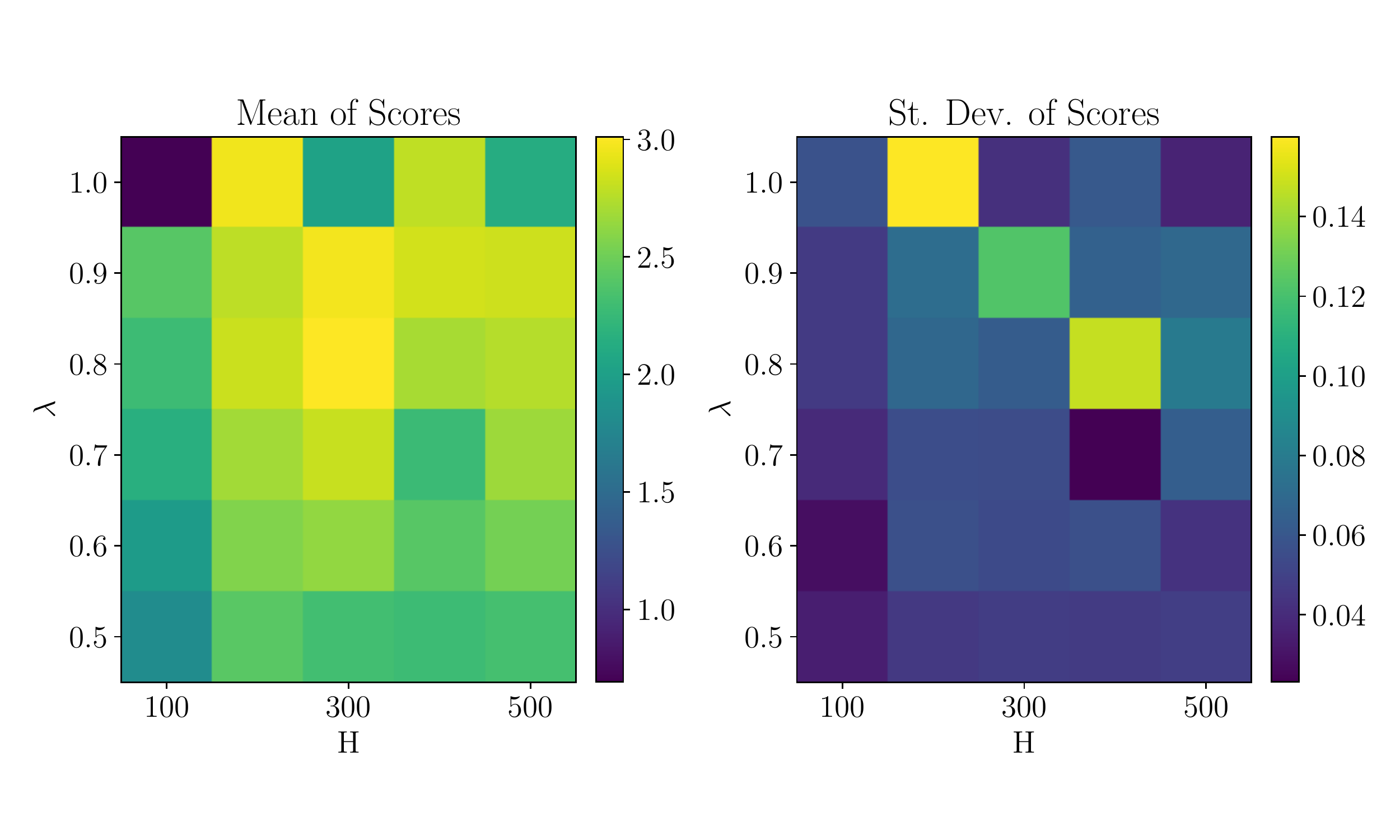}
\caption[]{Mean validation scores, including standard deviations, of different neural networks for the 10-dimensional PK model. Tested were neural networks with one hidden layer of different sizes $H \in \{100, 200, 300, 400, 500\}$ and a multiplicative learning rate scheduler of multiplier $\lambda \in \{0.5, 0.6, 0.7, 0.8, 0.9, 1.0\}$.}
\label{fig:pharma_D10_hyper}
\end{figure}

We here wish to select the best neural network architecture for the task of finding $D=10$ optimal designs for the PK model. We test one-layer neural networks with a ReLU activation function after the input layer and hidden units $H \in \{100, 200, 300, 400, 500\}$. Furthermore, we also compare different multipliers $\lambda \in \{0.5, 0.6, 0.7, 0.8, 0.9, 1.0\}$ in a multiplicative learning rate schedule; essentially, this means that the initial learning rates of $l_{\psib}=10^{-3}$ and $l_d=10^{-2}$ are multiplied by $\lambda$ every $5{,}000$ epochs up until a maximum of $50{,}000$ training epochs. Using prior samples we then generate a validation set of size $50{,}000$ at $\dbf^\ast$ and use this to compute the MI lower bound, i.e.~the validation score, given the trained neural network with a certain hyper-parameter combination. Doing this for a range of $H$ and $\lambda$ yields the comparison of validation scores shown in Figure~\ref{fig:pharma_D10_hyper}. In this figure, we show the mean and standard deviation of validation scores computed with several validation sets, as we can generate synthetic data at will. The overall highest mean validation score is achieved by a neural network with $H=300$ and $\lambda=0.8$.

\subsection{Convergence of Designs}

\begin{figure}[!t]
\includegraphics[width=\linewidth]{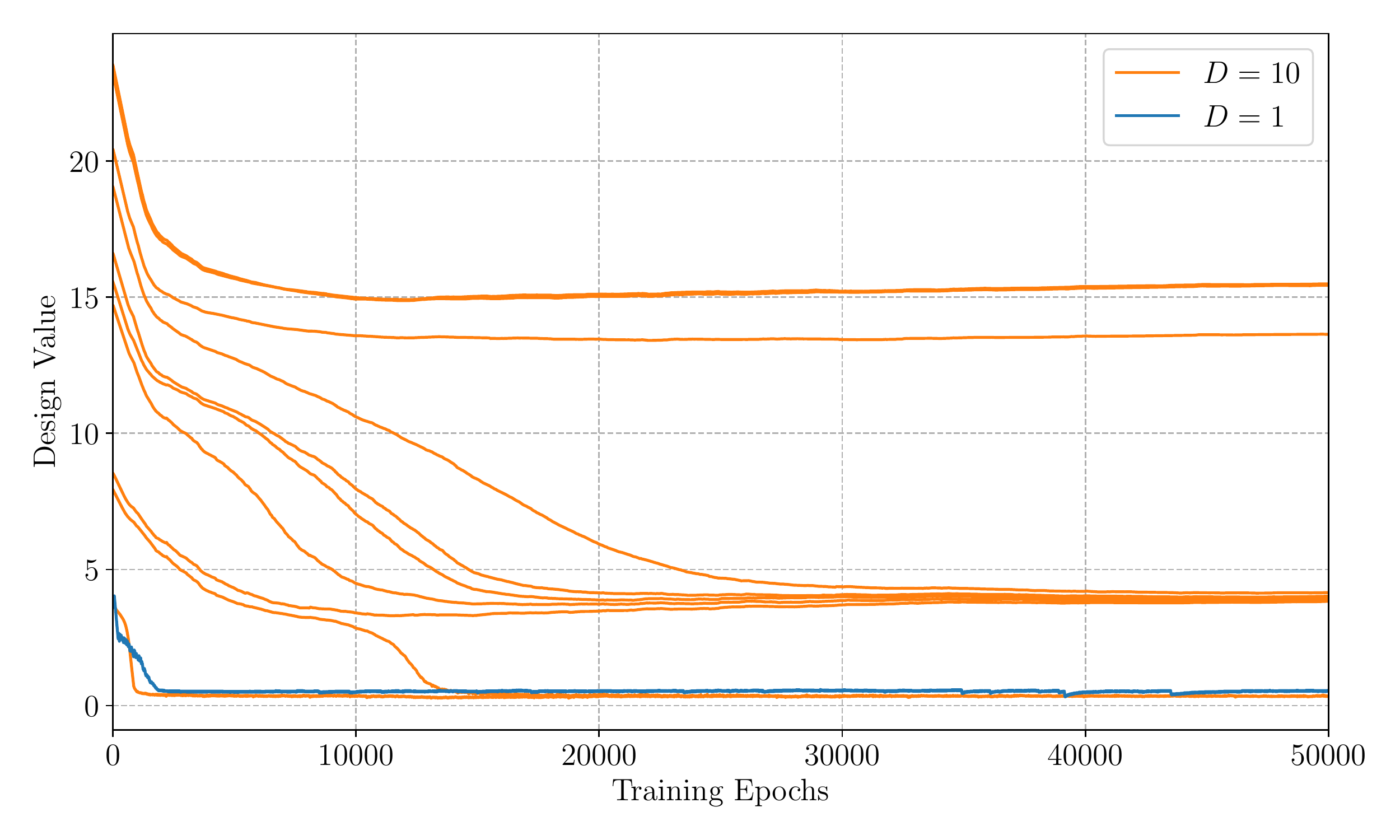}
\caption[]{Convergence of the individual design dimensions for the one-dimensional (blue curve) and 10-dimensional (orange curves) PK model. Note how the optimal design clusters for the 10-dimensional PK Model are spread over early, middle and late measurement times.}
\label{fig:pharma_convergence}
\end{figure}

In Figure~\ref{fig:pharma_convergence} we show the convergence of the
design vector for the one-dimensional (blue curve) and 10-dimensional
(orange curve) PK model. The one-dimensional design converges after
around $2{,}000$ training epochs, while the elements of the
10-dimensional design vector converge after roughly $30{,}000$
training epochs. The one-dimensional design ends up at a low design
time. Looking at the sampling path of the PK model and doing a Taylor
expansion for the exponents shows that this optimal design effectively
removes the effect of the elimination rate $k_e$ and estimates the
ratio $k_a / V$. For the 10-dimensional PK model, we have optimal
designs at early, middle and late measurement times. Late measurements
allow us to reduce the effect of $k_a$ and middle measurements are
needed to identify the remaining parameter. These optimal design
clusters are intuitive and match the ones found by~\citet{Ryan2014}.

\subsection{Full Posterior Plots}

\begin{figure}[!t]
\includegraphics[width=\linewidth]{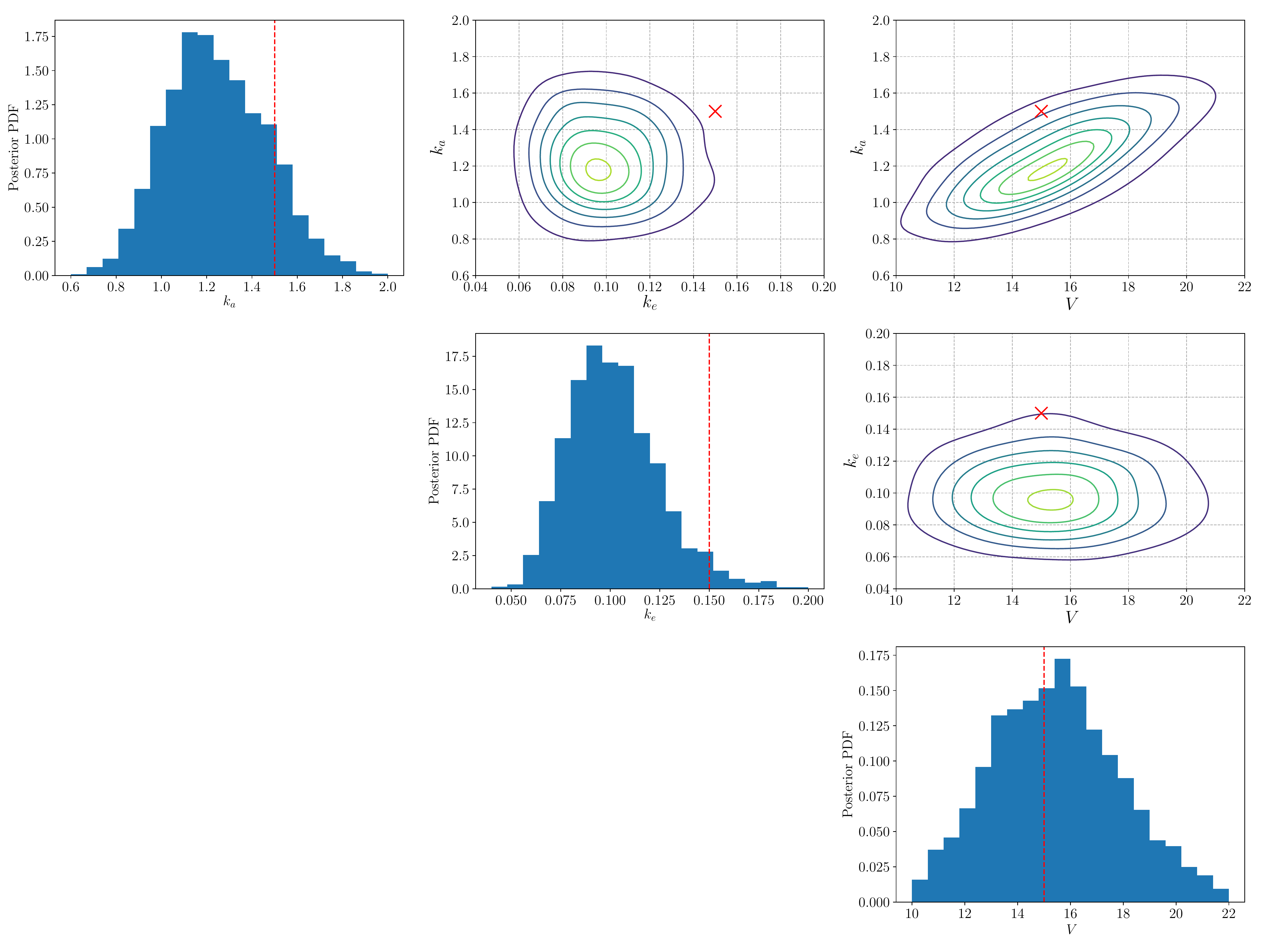}
\caption[]{Joint and marginal posterior distributions of the model parameters for the one-dimensional PK model, computed using posterior samples. Shown as red-dotted lines and red crosses are the true model parameters.}
\label{fig:pharma_post_D1_full}
\end{figure}

\begin{figure}[!t]
\includegraphics[width=\linewidth]{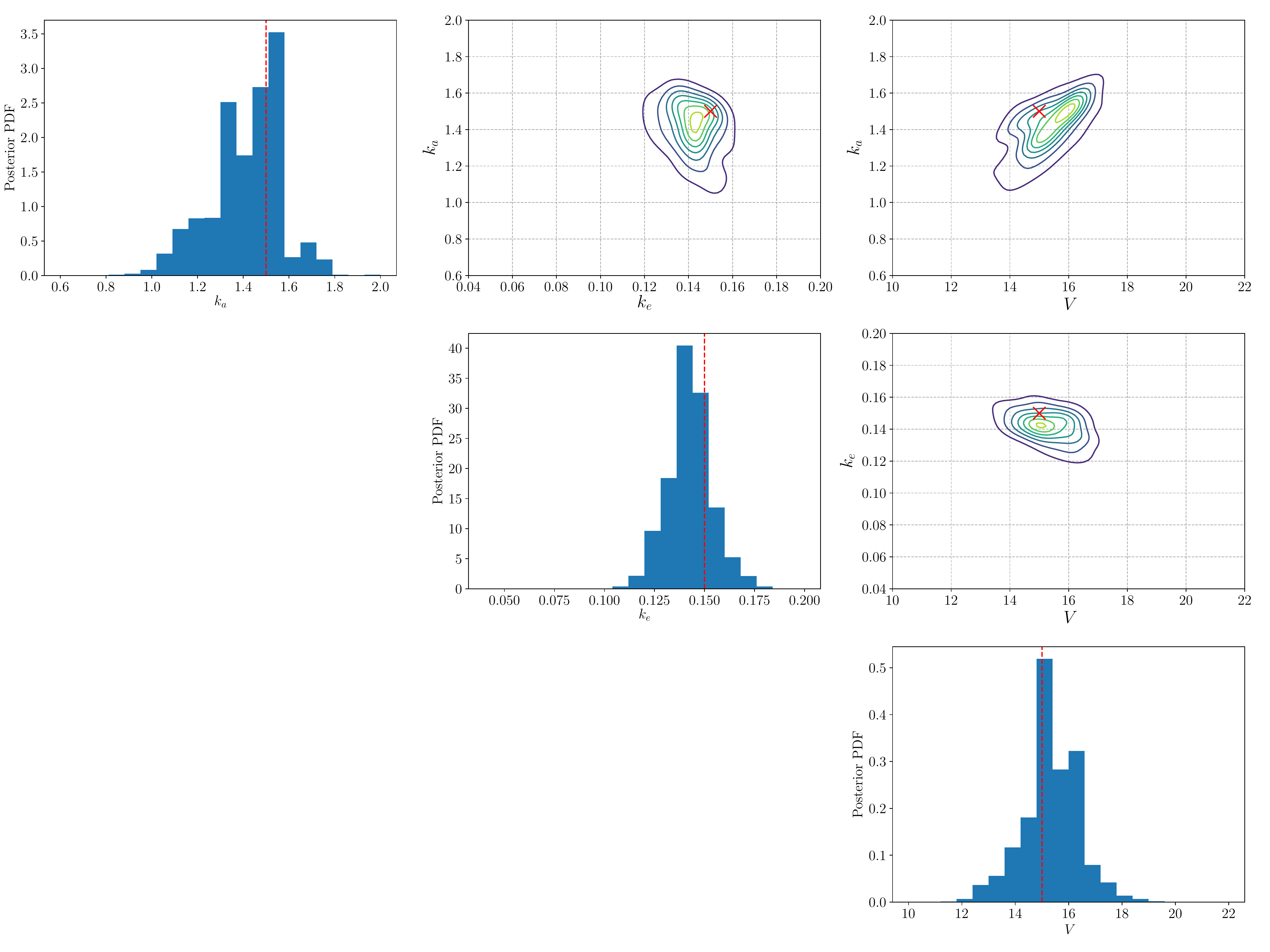}
\caption[]{Joint and marginal posterior distributions of the model parameters for the 10-dimensional PK model, computed using posterior samples. Shown as red-dotted lines and red crosses are the true model parameters.}
\label{fig:pharma_post_D10_full}
\end{figure}

We show the joint and marginal posterior distributions of the model parameters $\thetab=[k_a, k_e, V]^\top$ for the one-dimensional ($D=1$) and 10-dimensional ($D=10$) PK model in Figures~\ref{fig:pharma_post_D1_full} and~\ref{fig:pharma_post_D10_full}, respectively. As opposed to the one-dimensional PK model, the marginal posterior distributions for the 10-dimensional PK model are narrower and closer to the true model parameter values of $\thetab_{\text{true}} = [1.5, 0.15, 15]^\top$. Similarly, the mode of the joint posteriors for $D=10$ are closer to the $\thetab_{\text{true}}$ than for $D=1$. Interestingly, $k_a$ and $V$ are correlated for both $D=1$ and $D=10$, allowing us to measure the ratio of $k_a / V$ relatively well. 

\section{Additional Information for the Gas Leak Model}

\subsection{Hyperparameter Optimisation}

\begin{figure}[!t]
\includegraphics[width=\linewidth]{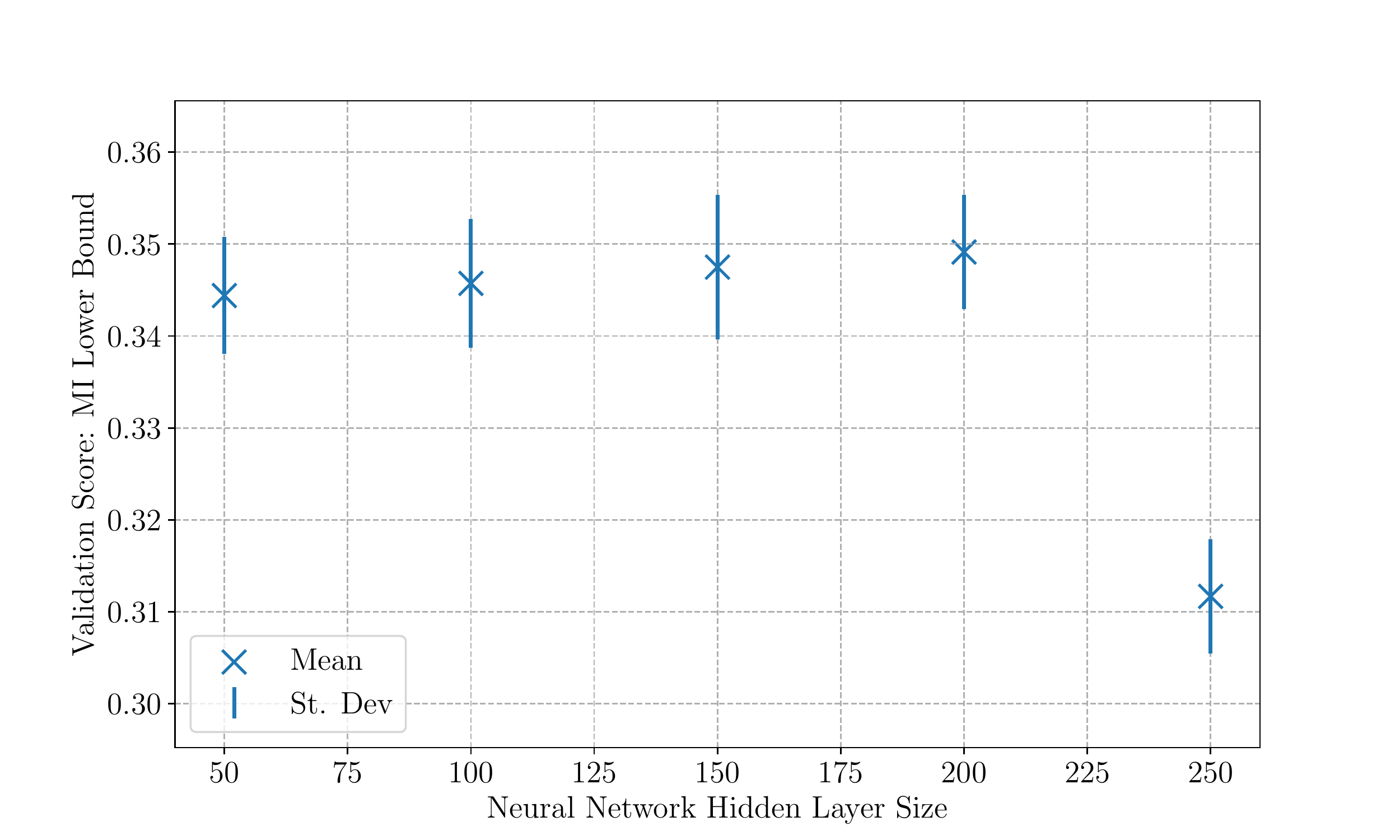}
\caption[]{Mean and standard deviation of validation scores (MI lower bound) for the Gas Leak model with $D=5$, using one-layer neural networks with different number of hidden units.}
\label{fig:gasleak_hyper}
\end{figure}

We here aim to find the optimal neural network size for the gas leak model when the number of measurements is $D=5$. To do so, we train several neural networks with one hidden layer of sizes $H \in \{50,100,150,200,250\}$ for $50{,}000$ epochs using learning rates $l_\psi = 10^{-3}$ and $l_d=10^{-2}$. By generating samples from the prior and the likelihood at the optimal design, we can build a validation set that we use to obtain an estimate of the MI lower bound. Repeating this several times for every trained neural network yields mean MI lower bound estimates, as well as standard deviations, as shown in Figure~\ref{fig:gasleak_hyper}. The highest validation score is achieved by a neural network with $H=200$.

\subsection{Additional Plots}

\begin{figure}[!t]
\includegraphics[width=\linewidth]{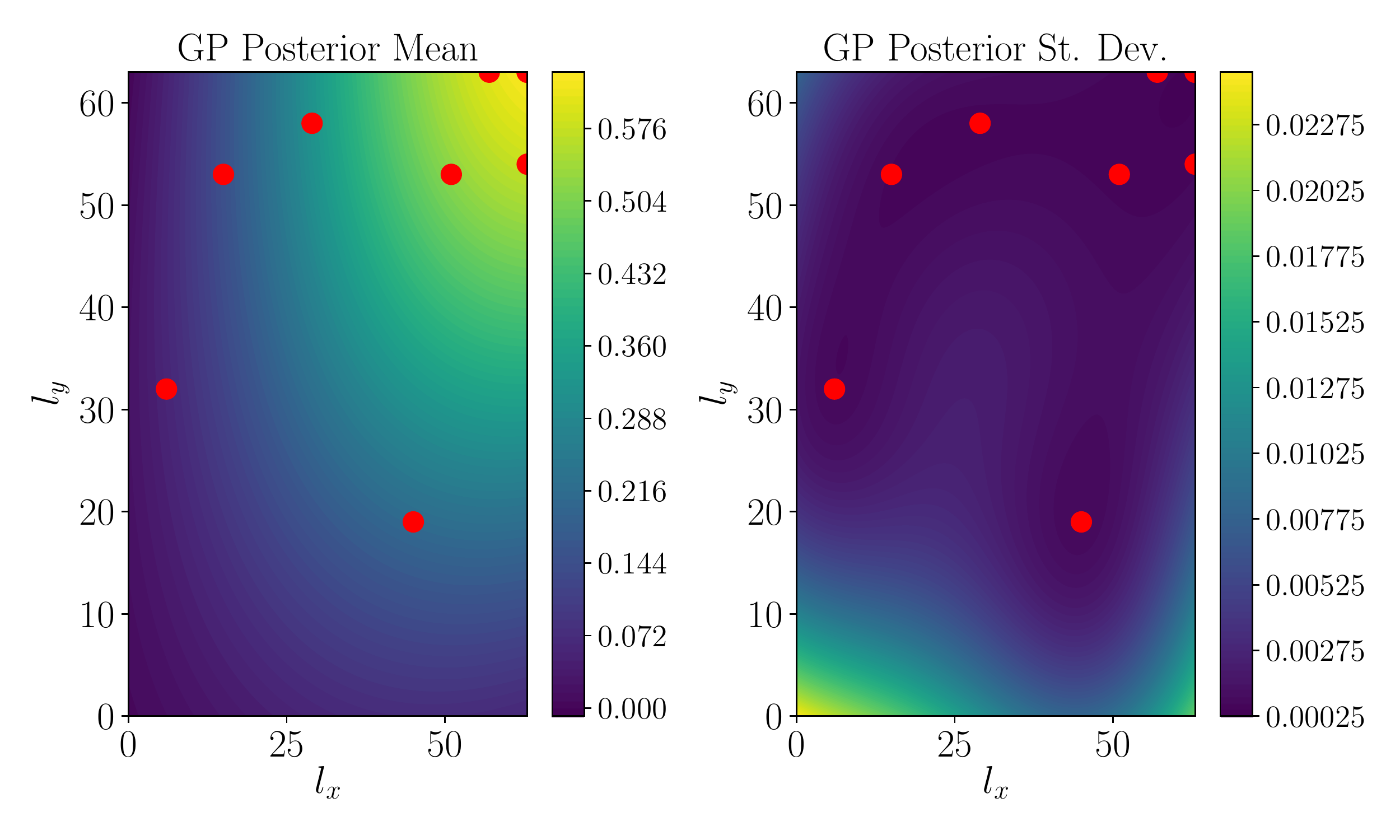}
\caption[]{GP posterior mean (left) and standard deviation of the MI lower bound surface for the gas leak model with $D=1$. Shown as red circles are the BO evaluations of the MI lower bound.}
\label{fig:gasleak_gp}
\end{figure}

To further visualise the design optimisation procedure when gradients of the sampling path are unavailable, we show the GP posterior mean of $\bound$ for the gas leak model when we perform $D=1$ measurement in Figure~\ref{fig:gasleak_gp}. The BO evaluations occur in locations where the MI lower bound is high, quickly converging to the optimum in the north east corner of the grid.

\begin{figure}[!t]
\includegraphics[width=\linewidth]{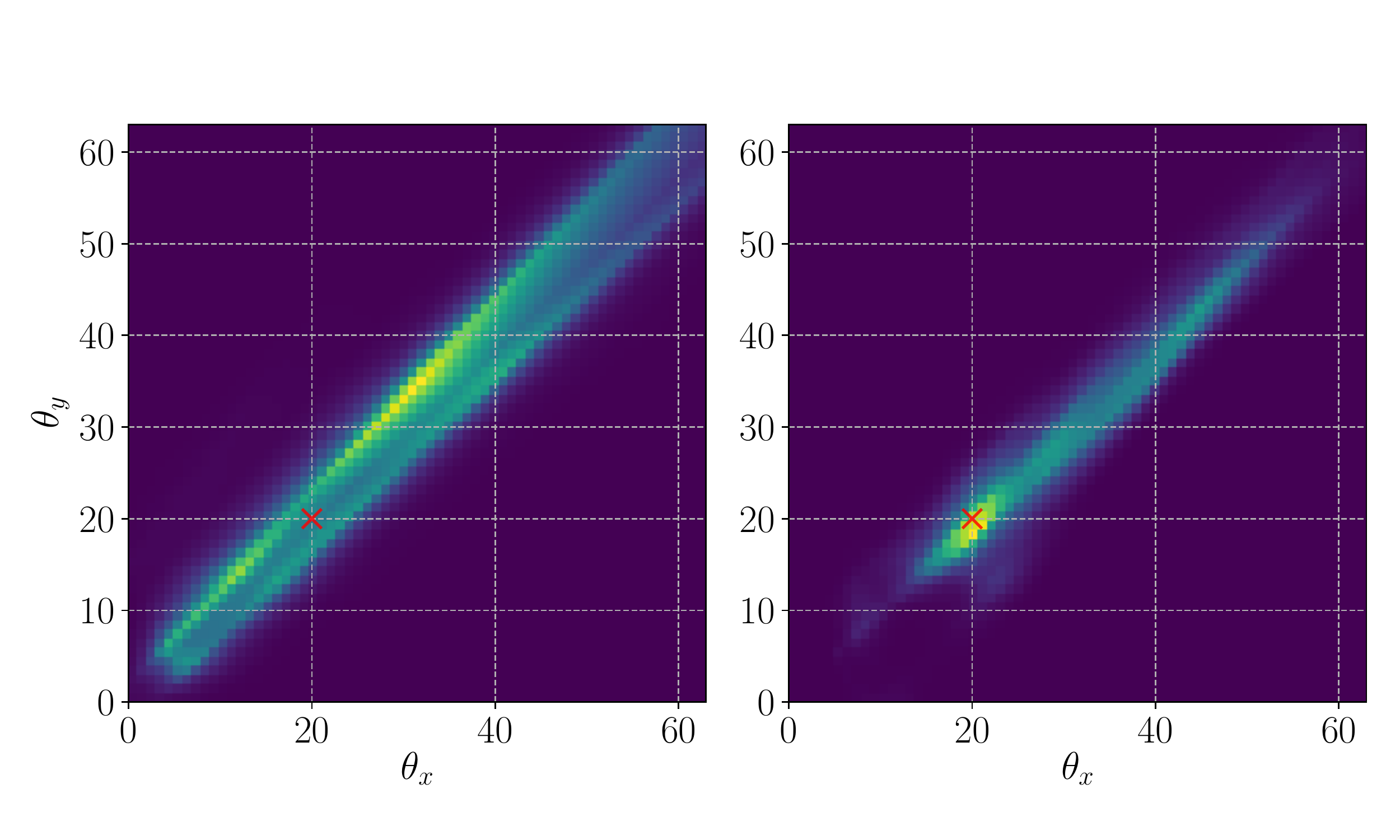}
\caption[]{Posterior density for the source locations of the gas leak with one measurement (left) and five measurements (right), assuming we know that the wind points in the direction of $W_d = 45^\circ$. Shown as the red cross is the true gas leak location.}
\label{fig:gasleak_post_wind45}
\end{figure}

In Figure~\ref{fig:gasleak_post_wind45} we show the posterior density of the gas leak source location $\thetab$ when we know that the wind direction is $W_d = 45^\circ$, for both $D=1$ (left) and $D=5$ (right). As opposed to the situation where we marginalised out the wind direction, see the main text, the posterior for $D=5$ is unimodal.

\section{Comparison with Bayesian D-Optimality}

Different utility functions in Bayesian experimental design tend to be geared towards different purposes. This usually makes a meaningful, direct comparison between utility functions difficult. Mutual information, which we have considered in this work, is used in order to optimally estimate model parameters. Another popular utility function that is used for parameter estimation is the Bayesian D-Optimality (BD-Opt),
\begin{equation} \label{eq:bdopt}
U(\dbf) = \mathbb{E}_{p(\ybf \mid \dbf)}\left[\frac{1}{\text{det}(\text{cov}(\thetab \mid \ybf, \dbf))}\right],
\end{equation}
which is a measure of how precise, on average, the resulting posterior is~\citep{Ryan2016}. We here provide a short comparison of optimal designs obtained via mutual information and BD-Opt.

We consider an oscillatory toy model that describes noisy measurements of a stationary waveform $\sin(\omega t)$. The design variable is the measurement time t and our experimental aim is to estimate the frequency $\omega$ in an optimal manner. The data-generating distribution is given by
\begin{align}
p(y \mid \omega, t) = \mathcal{N}(y; \sin(\omega t), \sigma_{\text{noise}}^2), \label{eq:sinean}
\end{align}
where $\sigma_{\text{noise}} = 0.1$ is the standard deviation of some measurement noise not depending on $t$; we here set the true model parameter to $\omega_{\text{true}}=0.5$. Furthermore, we choose a uniform prior distribution $p(\omega) = \mathcal{U}(\omega; 0, \pi)$ over the model parameter $\omega$. We note that reference posterior densities can be obtained by using the likelihood in~\eqref{eq:sinean} and Bayes' rule. This kind of model has also been considered by \citet{Kleinegesse2020} to illustrate their sequential Bayesian experimental design method.

We estimate and optimise the mutual information utility with our MINEBED framework, using a two-layered neural network with 100 hidden units each. The neural network is trained with the Adam optimiser and $10{,}000$ samples as the training set. The initial learning rates are $l_{\psi}=5\times10^{-3}$ and $l_{\mathbf{d}}=10^{-3}$, both multiplied by a factor of $0.9$ every $1{,}000$ epochs.

In order to estimate the BD-Opt utility in~\eqref{eq:bdopt} we require
samples from the posterior distribution (which we assume is
intractable for now). We here utilise LFIRE~\citep{Thomas2016} to
estimate the posterior density for a set of prior parameter samples. 
Using the posterior density and prior sample pairs, we then use 
categorical sampling to generate posterior samples
~\citep[see e.g.][]{Kleinegesse2019}. These
samples can then be used to approximate the determinant of the
covariance matrix needed in~\eqref{eq:bdopt}, for a given marginal
sample $y \mid t$. We can then approximate~\eqref{eq:bdopt} with a
Monte-Carlo sample average, using $1{,}000$ samples from $p(y|t)$.
We optimise the approximated BD-Opt utility by means of Bayesian
Optimisation~\citep{Shahriari2016} with a Gaussian Process surrogate
model.

Using our MINEBED framework we converge to an optimal design of
$t^\ast = 2.19$, whereas the optimum of the BD-Opt utility is $t^\ast
= 1.66$. We note that the time to converge to the optimum was
significantly lower for MINEBED than for the BD-Opt utility with
Bayesian Optimisation. While the optimal designs $t^\ast$ are quite
close, the values are still subtlety different, with BD-Opt favouring
smaller measurement times. This is because, by definition, BD-Opt
penalises posterior multi-modality that leads to larger variance
because it only takes into account the covariance matrix. Mutual
information on the other hand, is sensitive to multi-modality and
therefore does take into account multiple explanations for
observations.

\begin{figure}[!t]
\includegraphics[width=\linewidth]{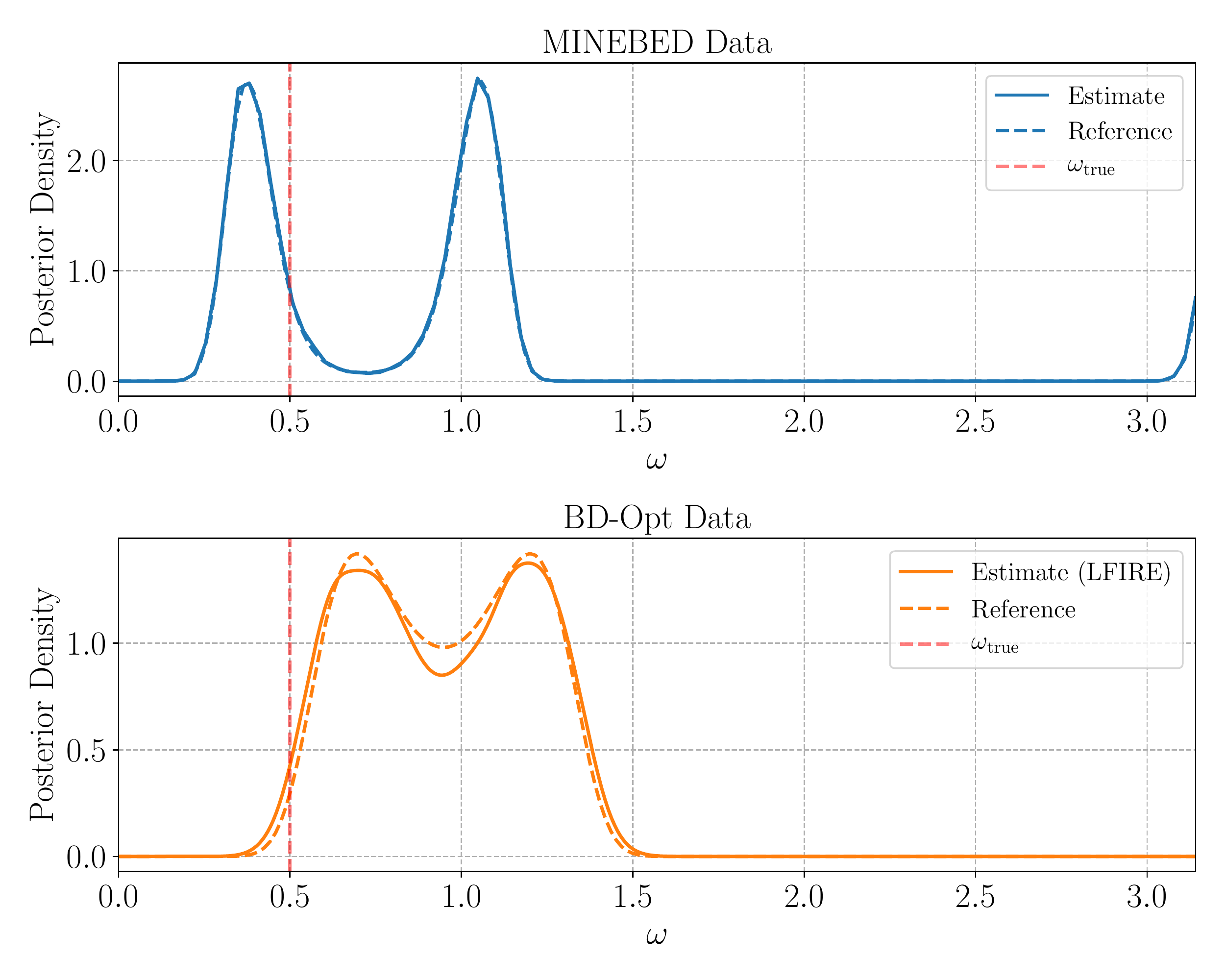}
\caption[]{Posterior densities for the oscillatory toy model using MINEBED data (top) and BD-Opt data (bottom). The dashed curves represent reference computations and the vertical dashed, red lines represent the true model parameter value.}
\label{fig:bdopt_comp}
\end{figure}

This is further reflected in Figure~\ref{fig:bdopt_comp}, where we
show the posterior densities for data obtained using MINEBED (top) and
BD-Opt (bottom). The real-world observations at $t^\ast$ were
generated using~\eqref{eq:sinean} and $\omega_{\text{true}}=0.5$. The
posterior density for MINEBED data was computed via the trained neural
network (see the main text), while for BD-Opt data this was done by
Gaussian kernel density estimation of the posterior samples. Reference 
posteriors are shown as dashed lines. Ultimately, the posterior obtained 
with BD-Opt data has modes that are closer together than for MINEBED 
data, as we would expect because it favours posteriors with small variances. 

\end{document}